\pdfoutput=1
\documentclass[11pt]{article}

\usepackage[]{acl}
\usepackage{times}
\usepackage{latexsym}
\usepackage{mathrsfs}
\usepackage{amsmath}
\usepackage{amssymb}
\usepackage{booktabs}
\usepackage{multirow}
\usepackage{graphicx}
\usepackage{multirow}
\usepackage{makecell}
\usepackage{hyperref}
\usepackage{adjustbox}
\usepackage{subcaption}

\usepackage[T1]{fontenc}
\usepackage[utf8]{inputenc}

\usepackage{microtype}

\DeclareMathOperator*{\argmax}{argmax}

\title{Language Models are Few-shot Multilingual Learners}
\author{Genta Indra Winata${^1\thanks{\hspace{2mm} Equal contribution}  }$  , Andrea Madotto$^{1,3*}$, Zhaojiang Lin$^{1}$, \\\textbf{Rosanne Liu$^{2,3}$, Jason Yosinski$^{3}$, Pascale Fung$^{1}$} \\
  $^1$The Hong Kong University of Science and Technology \\
  $^2$Google Brain \hspace{5mm} $^3$ML Collective \\
  \texttt{\{giwinata, amadotto, zlinao\}@connect.ust.hk}}

\begin{document}
\maketitle
\begin{abstract}
General-purpose language models have demonstrated impressive capabilities, performing on par with state-of-the-art approaches on a range of downstream natural language processing (NLP) tasks and benchmarks when inferring instructions from very few examples. Here, we evaluate the multilingual skills of the GPT and T5 models in conducting multi-class classification on non-English languages without any parameter updates. We show that, given a few English examples as context, pre-trained language models can predict not only English test samples but also non-English ones. Finally, we find the in-context few-shot cross-lingual prediction results of language models are significantly better than random prediction, and they are competitive compared to the existing state-of-the-art cross-lingual models and translation models. 
\end{abstract}

\section{Introduction}
The progress in language model (LM) pre-training~\cite{peters2018deep,devlin2019bert,radford2019language,yang2019xlnet,liu2019roberta,brown2020language,liu2020multilingual,lewis2020bart,raffel2020exploring,gao2020pile} has led to the possibility of conducting few-shot learning, that is, learning a new task using a small number of examples without any further training or gradient computation. Few-shot learning alleviates the cost for extensive labeled data, which is beneficial since collecting high-quality labeled data is resource-intensive and expensive. It also reduces the cost for model fine-tuning, which requires tremendous GPU or TPU resources. Few-shot learning can be seen as a \textit{one-for-all plug-and-play} computational model that can be applied to various natural language tasks, from sentiment analysis for text classification to story generation, provided only a small context~\cite{brown2020language}.

\begin{figure}[!th]
  \centering
  \includegraphics[width=\linewidth]{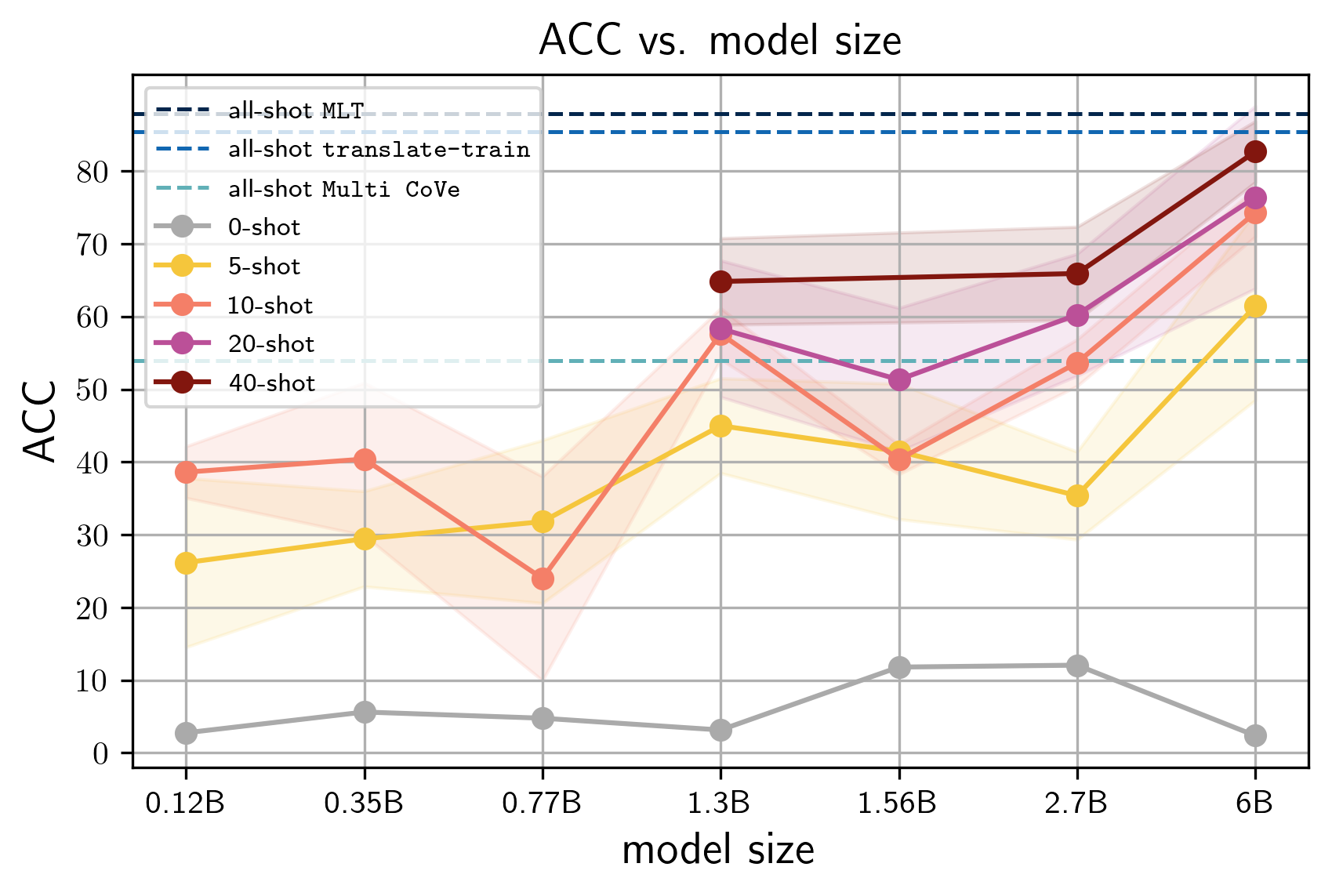} 
  \caption{The average accuracy vs. model size on English-Spanish Multilingual NLU dataset achieved by cross-lingual in-context learning using various GPT and T5 models. The shaded region represents the standard deviation of three runs. The all-shot results are taken from~\citet{liu2020attention}.}
  \label{fig:result}
\end{figure}

% In this paper, we investigate the few-shot learning ability under a multilingual setting. We show that, given few examples in English, pre-trained language models not only can few shot to English test samples, but also non-English ones. Different to previous zero-shot cross-lingual works~\cite{ponti2018adversarial,artetxe2019massively,liu2019zero,lauscher2020zero,liu2020attention,liu2021x2parser,chen2021zero} which trained models on all the examples in the source language, we study the more extreme case where only few examples in the source language are available, and we refer this setting as \textit{True Few Shot Cross-lingual}.
\begin{figure*}[!th]
    \centering
    \resizebox{\textwidth}{!}{  
    \includegraphics{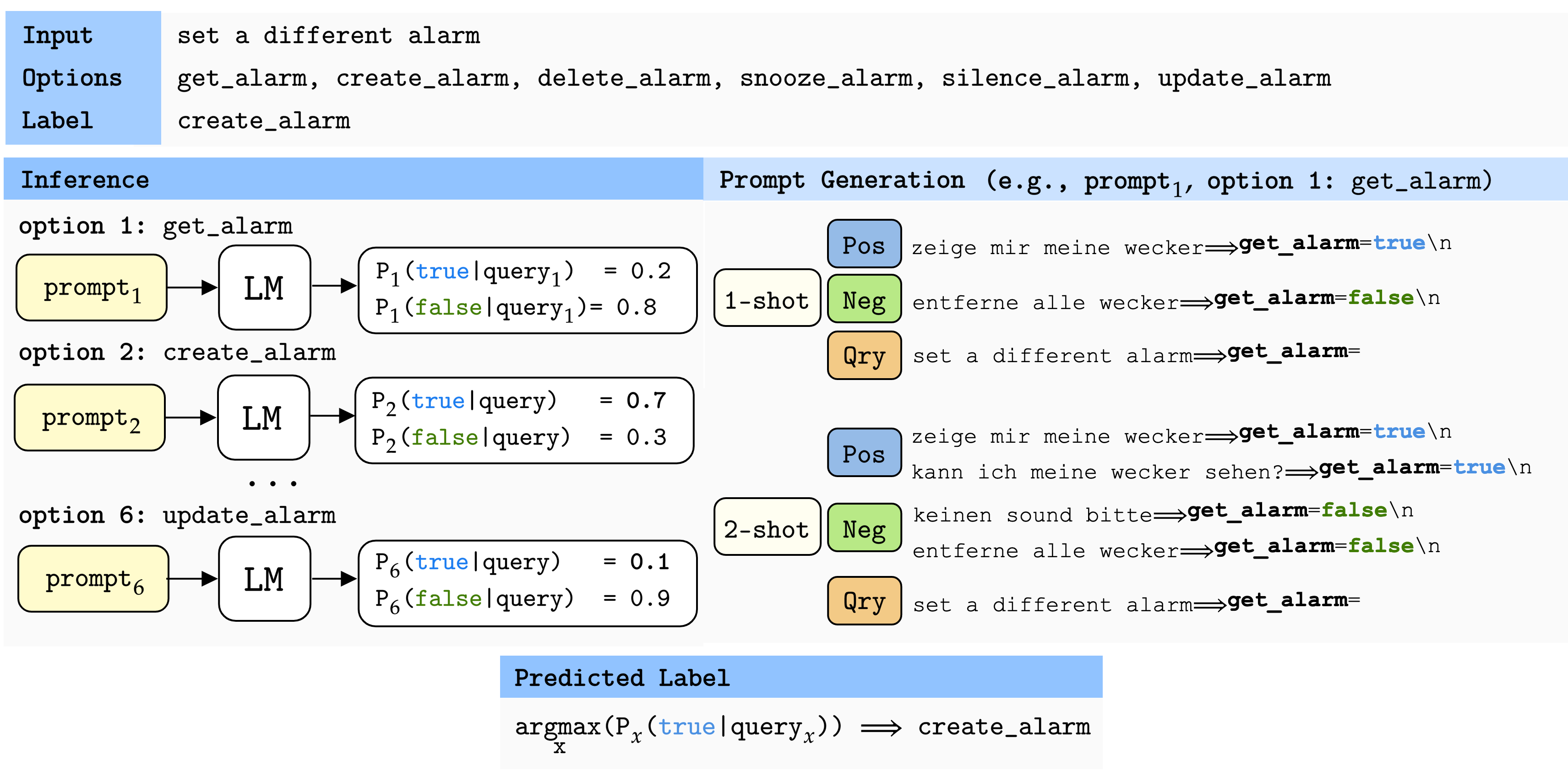}
    }
    \caption{Example of the inference and query generation on the few-shot learning, where the source language and target language are German and English, respectively.}
    \label{fig:flow}
\end{figure*}

The idea of few-shot learning is also relevant to address the low-resource issue in non-English languages. Few-shot learning has been applied to NLP tasks~\cite{brown2020language, madotto2020language,lu2021fantastically,perez2021true,liu2021makes,liu2021pre,cahyawijaya2021greenformer}. Common approaches to solve the low-resource issue are to pre-train models with self-supervised learning using unlabelled monolingual text data collected from various resources available online~\cite{wilie2020indonlu,le2020flaubert,martin2020camembert,eddine2020barthez,nguyen2020phobert,scheible2020gottbert,bhattacharjee2021banglabert,lee2020kr,cahyawijaya2021indonlg,park2021klue} and then apply pre-training on the source language and fine-tune on the target languages~\cite{schuster2019cross,lin2019choosing,winata2019code,winata2021multilingual,pfeiffer2020mad,zheng2021consistency,lin2021bitod}. Conversely, the few-shot learning does not need any training from the source and target languages. Figure~\ref{fig:result} shows how it is possible to utilize pre-trained models on non-English languages, such as Spanish, as the performance is not random, and the performance increases as the models are given more samples. We conjecture that pre-trained models may be able to adapt to languages that are similar to English. However, for many language tasks, it is difficult to collect a large supervised training dataset as language experts (e.g., linguists or native speakers) are required to annotate the data.

Another line of work is to apply cross-lingual transfer on English with the same task as the target languages~\cite{ponti2018adversarial,artetxe2019massively,liu2019zero,lauscher2020zero,liu2020attention,liu2021x2parser,chen2021zero}. However, such methods still need to apply a fine-tuning step to update the model for fast adaptation, which can be challenging for large pre-trained models -- some models require substantial memory capacity -- since the models have to be trained on high-performing machines. Different from the aforementioned method, in-context learning using a LM does not allow any parameter updates. Thus, the process does not need to compute and store the gradients for backward propagation. 

% Contributions
In this work, we investigate the practicality of applying few-shot learning in the multilingual setting for four languages, English, French, German, and Spanish, on natural language understanding intent prediction tasks using publicly available LMs that are mainly trained on English data. We show that, given a few English examples as context, pre-trained LMs can predict not only English test samples, but also non-English ones (Figure~\ref{fig:flow}). To the best of our knowledge, no existing works have studied these tasks in multilingual settings. We conjecture that the English LMs can still produce good results on languages that are closely related to English. We construct the inference for the multi-class prediction setup by extending the idea from~\citet{madotto2020language} of applying multiple binary predictions on each class. Instead of guiding the model to generate \textit{true} or \textit{false} like in their work, which is not consistent and sometimes generates other words --, we introduce \textit{maximum confidence prediction}. This method considers the confidence of predicting a certain label to provide a prediction. We design this as a multiple-choice task in which the confidence of the prediction for all possible classes is compared. Each class's confidence score is computed by normalizing the logits of generating the \textit{next boolean token} given the prompt as the context. This method is considered to be more scalable than the simple $k$-way few-shot learning, where we need to put all data in a single prompt, since we only have a fixed maximum sequence length and, in the deployment, each forward step can be run in parallel to speed up the process. To increase the difficulty of the challenge, we also propose a cross-lingual task, where the context and query are in different languages.

Overall, we find that conditional generative LMs, such as the GPT-2~\cite{radford2019language}, GPT$_{\text{NEO}}$ models~\cite{gao2020pile}, and T5 models~\cite{raffel2020exploring} have the capability to predict non-English languages, and adding more shots and using larger models achieves a substantial increment in performance, making it significantly better than random, which indicates the models are able to understand the prompt. We only focus on GPT and T5 models. T5 models do not perform as well as GPT models, which might be caused by the pre-training strategy. Experimental results in the cross-lingual setting demonstrate that pre-trained LMs make correct predictions. To summarize, our contributions are as follows:
\begin{itemize}
    \item We study few-shot learning in the multilingual setting on four languages without any gradient updates. We use the publicly available GPT and T5 LMs, and compare the results to those from the zero-shot and fine-tuning approaches.
    \item We propose a simple and straightforward approach to perform few-shot learning on multi-class classification by applying binary prediction and considering the confidence of predicting the boolean tokens.
    \item We display the zero-shot, one-shot, and many-shot proficiency of the LMs in the cross-lingual setting when the language of the prompt is different from the target language.
\end{itemize}

\section{Few-shot Multilingual Learners}
First, we briefly define the notation of the input and output of the task, and then we introduce our method to design prompts for few-shot in-context learning.
% \footnote{The code will be released after the submission to reproduce the results.}
\footnote{The code is released at \url{https://github.com/gentaiscool/few-shot-lm}.}

\subsection{Notation and Tasks}
Let us define $D$ as the distribution over the dataset and $P$ as the prompt that we use as the input of the LM $\theta$. The prompt $P=[D_{pos},D_{neg}, Q]$ is a concatenation of few-shot samples: positive samples $D_{pos}$, negative samples $D_{neg}$, and the query $Q$, where $D_{pos}$, $D_{neg}$ $\sim$ $D$. $D_{pos}$ is a sample with a label that is the same as the query, and $D_{neg}$ is a sample that is taken from the dataset $D$ with a label other than the query. $\theta$ takes $P$ as the input of the model, and the LM generates a word $y$. We define the task $T_{s \rightarrow t}$, where $s$ is the source language and $t$ is the target language. 

In this paper, we focus on the intent detection task in the monolingual and cross-lingual settings. In the monolingual setting, the source language is the same as the target language, and in the cross-lingual setting, we take the source language as different from the target language ($s \neq t$). We design our task as a multiple-choice problem, in which each sample has a label $l \in L$, where $L$ is the set of possible labels. We predict the boolean (true or false) for each sample and take the highest prediction confidence.

\subsection{Prompt Generation}
We define the task by designing prompts to perform few-shot learning. We design our task as a binary classification for multi-class prediction by following~\citet{madotto2020language}. The idea is to guide the model to predict the boolean tokens, \textit{true} and \textit{false}. We examine the usage of two types of LMs, GPT and T5 models, and we construct prompts specific to each model. We use a specific way to probe the LMs to perform the few-shot prediction since they are trained with different learning objectives. Table~\ref{prompt_format} shows the format of the prefix we use for the GPT and T5 models.
% \begin{align}
%     & \texttt{GPT} \nonumber \\
%     & \small{X_1 \rightarrow \texttt{true}, X_1^{*} \rightarrow \texttt{false} \cdots Q \rightarrow} \\
%     & \texttt{T5} \nonumber \\
%     & \small{X_1 \rightarrow \texttt{true}, X_1^{*} \rightarrow \texttt{false} \cdots Q \rightarrow \texttt{<MASK>}},
% \end{align}
\begin{table}[!ht] 
\centering
\resizebox{0.49\textwidth}{!}{ 
\begin{tabular}{ll} \toprule
Model & Prompt \\ \midrule
GPT & \texttt{[SAMPLES]} $Q$ $\rightarrow$ \\
T5 & \texttt{[SAMPLES]} $Q$ $\rightarrow$ \texttt{[MASK]} \\ \midrule \midrule
\multicolumn{2}{l}{\texttt{[SAMPLES]}}\\
Format & Example\\ \midrule
$X_1$ $\rightarrow$ \texttt{true\textbackslash n} & zeige mir meine wecker=>get\_alarm=true\textbackslash n\\ 
$X_1^{*}$ $\rightarrow$ \texttt{false\textbackslash n} & entferne alle wecker=>get\_alarm=false\textbackslash n\\
$\cdots$ & $\cdots$ \\
$X_k$ $\rightarrow$ \texttt{true\textbackslash n} & kann ich meine wecker sehen?=>get\_alarm=true\textbackslash n\\
$X_k^{*}$ $\rightarrow$ \texttt{false\textbackslash n} & keinen sound bitte=>get\_alarm=false\textbackslash n\\ \bottomrule
\end{tabular} 
}
\caption{Prompt format given a few German examples as context.}
\label{prompt_format}
\end{table}
$X_i$ is one of the few-shot samples, and $X_i^{*}$ is the sample from other classes. For the GPT models, we only input the prefix by concatenating positive and negative samples with the query. Specifically for the T5 models, we add an additional token after the query and let the model predict that particular token during the generation step. 

Figure~\ref{fig:flow} shows an example of how we generate the prompt in $k$-shot settings. We create $L$ prompts and apply $L$ forward steps for each sample. For each prompt, $k$ positive and negative samples are randomly drawn from the dataset. It is worthwhile to note that the sampling method is similar to $k$-way few-shot learning, but the samples are not merged into a single prompt. We do this because we want to give more shots as the prompt to the LMs as they have a limitation on the number of tokens they can accept as input (1,024 tokens in GPT-2$_{\text{XL}}$ and 2,048 tokens in GPT$_{\text{NEO}}$). We add a special token \texttt{\textbackslash n} as a separator between each sample, as shown in Table~\ref{prompt_format}.  

\subsection{Maximum Confidence Prediction}
To get the final prediction of each sample, first, we compute the score of predicting the next boolean (true or false) given the prompt $X_i$ for label $i$: $P_{\theta}(y=\texttt{true}|X_i)$ and $P_{\theta}(y=\texttt{false}|X_i)$ from the prediction distribution. Then, we normalize the score to get the probability of generating the \textit{true} token to measure how much confidence the LM has to predict label $i$. We collect all the confidence scores over all label options and choose the highest confidence score among them, as follows:
\begin{align}
\text{MC}(X, L) = \argmax_{i \in L}\frac{P_{\theta}(y=\texttt{true}|X_i)}{\sum_{b} {P_{\theta}(y=b|X_i)}},
\end{align}
where $b \in \{\texttt{true}, \texttt{false}\}$. We take the label with the highest confidence score as $\text{MC}(X, L)$.

\subsection{Choices of Samples}
For in-context learning, choosing the order of samples is essential~\cite{lu2021fantastically}. Here, we examine the impact of the order of the samples. We construct the probing set in two ways: (1) shuffle the few-shot samples and measure the variance in performance after changing their order, and (2) arrange the positive samples before the negative samples. We find that the latter works well, specifically on the T5 models.

\section{Baselines}
In this work, we compare the few-shot learning performance with other common approaches: zero-shot, zero-shot cross-task, and fine-tuning. 
\subsection{Zero-shot Cross-Task}
One way to solve zero-shot prediction is by using entailment models to calculate the entailment score between sequences and labels. Given a pre-trained LM $\psi$ with an entailment head, a set of hypotheses $H$, and possible labels $L$, the model accepts two inputs, the hypothesis $h \in H$ and label $l \in L$, and generates the entailment score given any combinations of the hypothesis and label $P_{\psi}(y=\texttt{entail}|h,l)$: 
\begin{align}
    \text{ES}(H, L) = \argmax_{h,l\in \{H,L\}}{P_{\psi}(y=\texttt{\small{entail}}|h,l)}.
\end{align}

\subsection{Zero-shot In-Context Learning}
This approach is very similar to our few-shot approach. It does not need any samples, and the model is only given natural language instruction. However, instead of using the prompt like in the few-shot setting, we can set up the prompt in a question-and-answer (Q\&A) format as follows:
\begin{align}
&\texttt{Q:}\text{\small{ Is \texttt{`<INTENT>'} the intent of \texttt{`<TEXT>'}}? } \texttt{A:}.
\end{align}

\begin{table*}[!th]
\centering
\resizebox{\linewidth}{!}{
\begin{tabular}{l|c|cccc|cc}
\toprule
\multicolumn{1}{l}{\multirow{2}{*}{\textbf{Models}}} & \multicolumn{1}{c}{\multirow{1}{*}{\textbf{SNIPS}}} & \multicolumn{4}{c}{\textbf{MTOP}} & \multicolumn{2}{c}{\textbf{MultiNLU}} \\ \cmidrule(l{2pt}r{2pt}){2-2} \cmidrule(l{2pt}r{2pt}){3-6} \cmidrule(l{2pt}r{2pt}){7-8}
\multicolumn{1}{c}{} & \multicolumn{1}{c}{\textbf{en}} & \multicolumn{1}{c}{\textbf{de}} & \multicolumn{1}{c}{\textbf{en}} & \multicolumn{1}{c}{\textbf{es}} & \multicolumn{1}{c}{\textbf{fr}} & \multicolumn{1}{c}{\textbf{en}} & \multicolumn{1}{c}{\textbf{es}} \\ \midrule
% Majority (Training set) & 12.29 & 67.37 & 68.35 & 69.38 & 65.80 & 45.84 & 33.68 \\ 
Random & 14.29 & 15.07 & 15.25 & 15.55 & 14.36 & 8.33 & 8.33 \\
Full-training SOTA & 99.00$^\ddagger$ & 88.80$^\dagger$ & 94.00$^\dagger$ & 90.10$^\dagger$ & 89.60$^\dagger$ & 99.11$^*$ & 98.90$^*$ \\ \midrule
\multicolumn{8}{c}{\textit{Zero-shot Cross-Task Prediction}} \\ \midrule
BART$_{\text{LARGE}}$ \texttt{0.4B} & 74.43 & 24.80 & 43.41 & 36.06 & 24.77 & 65.60 & 34.77 \\
XLM-R$_{\text{LARGE}}$ \texttt{0.6B} & 68.00 & 54.30 & 53.37 & 51.67 & 51.99 & 77.79 & 66.35 \\ \midrule
\multicolumn{8}{c}{\textit{Few-shot Learning (K-shot)}} \\ \midrule
GPT-2 \texttt{0.1B} & 39.33 $\pm$ 8.58 & 40.03 $\pm$ 6.34 & 35.46 $\pm$ 0.92 & 36.18 $\pm$ 2.12 & 41.16 $\pm$ 5.65 & 51.59 $\pm$ 12.83 & 37.56 $\pm$ 7.14 \\
GPT-2$_{\text{MEDIUM}}$ \texttt{0.3B} & 65.71 $\pm$ 2.80 & 52.94 $\pm$ 5.12 & 63.35 $\pm$ 3.01 & 54.33 $\pm$ 4.75 & 50.6 $\pm$ 2.44 & 72.21 $\pm$ 14.88 & 50.25 $\pm$ 4.99 \\
GPT-2$_{\text{LARGE}}$ \texttt{0.8B} & 71.43 $\pm$ 10.27 & 50.94 $\pm$ 6.63 & 59.70 $\pm$ 4.50 & 52.38 $\pm$ 2.65 & 44.75 $\pm$ 1.11 & 62.36 $\pm$ 13.82 & 58.04 $\pm$ 5.28 \\
GPT-2$_{\text{XL}}$ \texttt{1.6B} & 78.43 $\pm$ 3.16 & 78.43 $\pm$ 3.16 & 73.93 $\pm$ 1.21 & 56.61 $\pm$ 2.02 & 45.21 $\pm$ 2.54 & 79.04 $\pm$ 5.05 & 64.74 $\pm$ 7.64 \\ 
GPT$_{\text{NEO}}$ \texttt{1.3B} & 84.19 $\pm$ 2.78 & 67.17 $\pm$ 2.50 & 82.40 $\pm$ 1.90 & 73.51 $\pm$ 0.95 & 66.3 $\pm$ 1.29 & 89.70 $\pm$ 1.28 & 85.77 $\pm$ 2.53 \\
GPT$_{\text{NEO}}$ \texttt{2.7B} & 91.24 $\pm$ 0.68 & 71.57 $\pm$ 5.94 & 81.51 $\pm$ 0.39 & 76.94 $\pm$ 0.83 & 70.31 $\pm$ 1.99 & 83.76 $\pm$ 3.14 & 87.82 $\pm$ 1.55 \\
GPT$_{\text{NEO-J}}$ \texttt{6B} & \textbf{93.38 $\pm$ 0.76} & \textbf{80.97 $\pm$ 3.21} & \textbf{89.66 $\pm$ 0.50} & \textbf{84.18 $\pm$ 0.32} & \textbf{85.04 $\pm$ 1.18} & \textbf{94.32 $\pm$ 1.14} & \textbf{88.54 $\pm$ 6.18} \\ 
T5$_{\text{LARGE}}$ \texttt{0.8B} & 23.57 $\pm$ 8.93 & 41.84 $\pm$ 7.63 & 36.02 $\pm$ 5.26 & 49.49 $\pm$ 6.32 & 40.41 $\pm$ 5.97 & 37.57 $\pm$ 15.23 & 21.20 $\pm$ 6.51 \\
T5$_{\text{3B}}$ \texttt{3B} & 46.52 $\pm$ 6.69 & 50.81 $\pm$ 6.45 & 46.17 $\pm$ 4.06 & 46.45 $\pm$ 4.39 & 44.38 $\pm$ 0.22 & 31.46 $\pm$ 18.18 & 31.60 $\pm$ 14.90 \\ \midrule
GPT$_{\text{NEO}}$ \texttt{2.7B} (ordered) & 86.71 $\pm$ 1.62 & 55.69 $\pm$ 3.45 & 55.12 $\pm$ 4.01 & 50.77 $\pm$ 4.41 & 50.70 $\pm$ 2.47 & 63.33 $\pm$ 7.14 & 61.51 $\pm$ 1.63 \\
T5$_{\text{LARGE}}$ \texttt{0.8B} (ordered) & 25.90 $\pm$ 18.51 & 63.06 $\pm$ 4.56 & 51.92 $\pm$ 3.90 & 62.71 $\pm$ 6.30 & 55.91 $\pm$ 3.82 & 38.97 $\pm$ 14.80 & 63.10 $\pm$ 4.46 \\
T5$_{\text{3B}}$ \texttt{3B} (ordered) & \textbf{93.00 $\pm$ 3.00} & \textbf{74.11 $\pm$ 2.69} & \textbf{65.03 $\pm$ 1.87} & \textbf{66.97 $\pm$ 1.35} & \textbf{68.89 $\pm$ 2.51} & \textbf{80.12 $\pm$ 3.95} & \textbf{86.60 $\pm$ 2.40} \\ \midrule
\multicolumn{8}{c}{\textit{Fine-tuning (40-shot)}} \\ \midrule 
mBERT \texttt{0.2B} & 88.57 $\pm$ 3.14 & 25.21 $\pm$ 2.31 & 41.44 $\pm$ 5.59 & 33.82 $\pm$ 10.08 & 16.54 $\pm$ 5.54 & 84.88 $\pm$ 1.59 & 87.87 $\pm$ 3.29 \\
XLM-R$_{\text{BASE}}$ \texttt{0.3B} & 87.95 $\pm$ 1.39 & 27.47 $\pm$ 11.90 & 37.03 $\pm$ 5.11 & 27.16 $\pm$ 5.51 & 13.8 $\pm$ 6.50 & 77.06 $\pm$ 3.16 & 74.85 $\pm$ 1.53 \\ \bottomrule
\end{tabular}
}
\caption{Zero-shot and few-shot results in the monolingual setting. The SOTA results are taken from $^\dagger$\citet{li2021mtop}, $^\ddagger$\citet{qin2019stack}, and $^*$\citet{schuster2019cross}.}
\label{monolingual-results}
\end{table*}

\subsection{Fine-tuning}
Fine-tuning is the most common approach to updating a pre-trained model's weights when training with a labeled dataset. The advantage of this approach is strong performance since we give supervised signals with the correct labels to the model. For fine-tuning, we use the same sets of few-shot samples as in the in-context learning. In Section~\ref{sec:Experiment}, we provide the hyper-parameters used in the experiments. 
% We apply an early stopping to find the best model on the validation set.
\section{Experiments}

\subsection{Datasets and Metrics}
We use an English natural language understanding (NLU) dataset, SNIPS~\cite{coucke2018snips}, and two multilingual NLU datasets, MTOP~\cite{li2021mtop} and Multilingual NLU (MultiNLU)~\cite{schuster2019cross}. MTOP includes four languages, English (\texttt{en}), French (\texttt{fr}), German (\texttt{de}), and Spanish (\texttt{es}), and Multilingual NLU includes two languages, English (\texttt{en}) and Spanish (\texttt{es}). We measure the model performance by calculating the average and standard deviation of the accuracy with three runs. 
% We measure the model performance by calculating the average and standard deviation of the accuracy and macro f1-score with three runs. 

\subsection{Experiment Settings}~\label{sec:Experiment}
We set up the experiment in two settings: monolingual and cross-lingual. In the monolingual setting, we test the ability of the model to conduct few-shot in-context learning on four languages: English (\texttt{en}), French (\texttt{fr}), German (\texttt{de}), and Spanish (\texttt{es}). In the cross-lingual setting, we test its ability to predict a query from a non-English language with the English context (\texttt{en$\rightarrow$XX}). In the few-shot in-context learning, we use $k$-way-few-shot classification, taking $k$ samples. For each model, we take $k\in[0,5,K]$, where $K \leq 40$ is the largest number of few-shot samples that can be passed to the model as input and is divisible by 10 without exceeding the maximum input token limit. We utilize an NVIDIA Tesla V100 16GB GPU to run the inference so that the model is ensured to fit in a single GPU, and we use 16-bit precision.

\paragraph{Model details} We run experiments on a variety of publicly available models:\footnote{The models except GPT$_{\text{NEO-J}}$ are taken from \url{https://huggingface.co/}. The GPT$_{\text{NEO-J}}$ model is taken from \url{https://github.com/kingoflolz/mesh-transformer-jax/}} four sizes of GPT-2 models (0.1B, 0.3B, 0.8B and 1.6B), three sizes of GPT$_\text{NEO}$ models (1.3B, 2.7B, and 6B), and two sizes of T5 models (0.8B and 3B). Table~\ref{model_list} shows the details of each pre-trained model.

\paragraph{Baselines} 
We use the same sets of few-shot samples for the baselines. We run fine-tuning on the pre-trained models mBERT~\cite{devlin2019bert} and XLM-R~\cite{conneau2020unsupervised}, and also compare our models with the zero-shot cross-task models using pre-trained models XLM-R, fine-tuned on XNLI~\cite{conneau2018xnli}, and BART, fine-tuned on MNLI~\cite{williams2018broad};\footnote{The XLM-R model fine-tuned with XNLI data can be accessed at \url{https://huggingface.co/joeddav/xlm-roberta-large-xnli}. The BART model fine-tuned with MNLI data can be accessed at \url{https://huggingface.co/facebook/bart-large-mnli}} a random baseline; and state-of-the-art results reported on each dataset. For the finetuning, we use a learning rate of 5e-5 with a decay of 0.9 for every epoch, and a batch size of 32. We apply an early stopping after 5 epochs without any improvement on the validation set.

\begin{table}[!th]
\centering
\resizebox{0.49\textwidth}{!}{ 
\begin{tabular}{lrccc} \toprule
Model Name & n$_{\text{params}}$ & n$_{\text{layers}}$ & n$_{\text{hidden}}$ & n$_{\text{ffn}}$ \\ \midrule
GPT-2 & 0.1B & 12 & 768 \\
GPT-2$_{\text{MEDIUM}}$ & 0.3B & 24 & 768 & - \\
GPT-2$_{\text{LARGE}}$ & 0.8B & 36 & 1,280 & - \\
GPT-2$_{\text{XL}}$ & 1.6B & 48 & 1,600 & - \\
GPT$_{\text{NEO}}$ & 1.3B & 24 & 2,048 & - \\
GPT$_{\text{NEO}}$ & 2.7B & 32 & 2,560 & - \\
GPT$_{\text{NEO-J}}$ & 6B & 28 & 4096 & 16,384 \\ \midrule
T5$_{\text{LARGE}}$ & 0.8B & 24 & 1,024 & 4,096 \\
T5$_{\text{3B}}$ & 3B & 24 & 1,024 & 16,384 \\
% T5$_{\text{11B}}$ & 11B & 24 & 1,024 & 65,536 \\
\bottomrule          
\end{tabular}
}
\caption{Model architecture.}
\label{model_list}
\end{table}

\begin{table*}[!th]
\centering
\resizebox{0.97\linewidth}{!}{
\begin{tabular}{l|ccc|c}
\toprule
\multicolumn{1}{l}{\multirow{2}{*}{\textbf{Models}}} &  \multicolumn{3}{c}{\textbf{MTOP}} & \multicolumn{1}{c}{\textbf{MultiNLU}} \\
\cmidrule(l{2pt}r{2pt}){2-4} \cmidrule(l{2pt}r{2pt}){5-5}
\multicolumn{1}{c}{} &  \multicolumn{1}{c}{\textbf{en$\rightarrow$de}} & \multicolumn{1}{c}{\textbf{en$\rightarrow$es}} & \multicolumn{1}{c}{\textbf{en$\rightarrow$fr}} & \multicolumn{1}{c}{\textbf{en$\rightarrow$es}} \\ \midrule
\multicolumn{5}{c}{\textit{Fine-tuning (all-shot on source language, zero-shot on target language)}} \\ \midrule
Seq2Seq w/ CRISS~\cite{li2021mtop} & 36.10 & 48.60 & 46.60 & - \\
Seq2Seq w/ XLM-R~\cite{li2021mtop} & 42.30 & 50.30 & 43.90 & - \\
NLM~\cite{liu-etal-2021-x2parser} & 54.91 & 59.99 & 58.16 & -\\
X2Parser~\cite{liu-etal-2021-x2parser} & \textbf{56.16} & \textbf{60.30} & \textbf{58.34} & -\\
Multi CoVe~\cite{schuster2019cross} & - & - & - & 53.89 \\ 
Translate-Train~\cite{liu2020attention} & - & - & - & 85.39 \\
MTL~\cite{liu2020attention} & - & - & - & \textbf{87.88} \\ \midrule
\multicolumn{5}{c}{\textit{Few-shot Learning (K-shot)}} \\ \midrule
GPT-2 \texttt{0.1B} & 23.89 $\pm$ 1.52 & 27.10 $\pm$ 3.19 & 26.14 $\pm$ 0.54 & 38.60 $\pm$ 3.54 \\
GPT-2$_{\text{MEDIUM}}$ \texttt{0.3B} & 39.61 $\pm$ 5.42 & 41.81 $\pm$ 4.66 & 42.40 $\pm$ 3.84 & 40.40 $\pm$ 10.48 \\
GPT-2$_{\text{LARGE}}$ \texttt{0.8B} & 30.94 $\pm$ 4.45 & 34.69 $\pm$ 6.50 & 33.04 $\pm$ 4.56 & 23.99 $\pm$ 14.02 \\
GPT-2$_{\text{XL}}$ \texttt{1.6B} & 42.88 $\pm$ 4.94 & 48.43 $\pm$ 4.42 & 50.67 $\pm$ 4.50 & 51.31 $\pm$ 9.87 \\ 
GPT$_{\text{NEO}}$ \texttt{1.3B} & 56.14 $\pm$ 2.75 & 63.14 $\pm$ 2.52 & 60.25 $\pm$ 3.32 & 64.82 $\pm$ 5.94 \\
GPT$_{\text{NEO}}$ \texttt{2.7B} & 58.27 $\pm$ 1.28 & 64.79 $\pm$ 1.69 & 62.30 $\pm$ 1.60 & 65.91 $\pm$ 6.42 \\
GPT$_{\text{NEO-J}}$ \texttt{6B} & \textbf{79.41 $\pm$ 1.18} & \textbf{81.57 $\pm$ 0.83} & \textbf{77.85 $\pm$ 1.63} & \textbf{82.66 $\pm$ 4.19} \\ 
T5$_{\text{LARGE}}$ \texttt{0.8B} & 37.14 $\pm$ 5.44 & 38.14 $\pm$ 3.20 & 33.53 $\pm$ 4.85 & 14.95 $\pm$ 16.34 \\
T5$_{\text{3B}}$ \texttt{3B} & 35.35 $\pm$ 7.07 & 34.64 $\pm$ 6.21 & 37.26 $\pm$ 8.68 & 14.11 $\pm$ 14.01 \\ \midrule
GPT$_{\text{NEO}}$ \texttt{2.7B} (ordered) \texttt{0.8B} & 42.23 $\pm$ 3.24 & 48.62 $\pm$ 2.60 & 46.30 $\pm$ 3.02 & 47.83 $\pm$ 5.73 \\
% T5$_{\text{LARGE}}$ (ordered) \texttt{0.8B} & x & x & x & 47.66 $\pm$ 5.29 \\
T5$_{\text{3B}}$ (ordered) \texttt{3B} & \textbf{52.23 $\pm$ 4.29} & \textbf{52.74 $\pm$ 3.20} & \textbf{49.72 $\pm$ 5.37} & \textbf{50.42 $\pm$ 6.01} \\ \bottomrule
\end{tabular}
}
\caption{Few-shot results in the cross-lingual setting on MTOP and MultiNLU datasets.}
\label{crosslingual-results}
\end{table*}

\section{Results and Analysis}

\subsection{Model Performance}
Tables~\ref{monolingual-results} and~\ref{crosslingual-results} show the results in the monolingual and cross-lingual settings, respectively. The tables show that the performance improvement is highly related to the size of the pre-trained model, and the performance gap between the fully trained state-of-the-art model and the few-shot learning models is decreasing when we use larger models, indicating the usefulness of utilizing models of bigger sizes. The performance of the models with few-shot learning is considered promising as they are not trained at all and the best model's performance gap with the fine-tuned model is less than 10\%.

\paragraph{Few-shot vs. Fine-tuning.}
Comparing the performance of generative models to fine-tuning, it is clear that we can achieve higher accuracy without any training. However, in this experiment, we acknowledge GPT and T5 models we use for in-context learning are larger than the models we fine-tune, and few-shot learning is much more efficient since the models are not required to store the intermediate memory. In terms of inference speed, the few-shot models require more time to run an inference step, which may cause a bottleneck when the number of few-shot samples is relatively large. This is the limitation of this method, and reducing the inference time is an open research area to improve the efficiency of in-context learning.

\paragraph{Zero-shot cross-task baselines.}
Surprisingly, the zero-shot cross-task models are able to predict the samples much better than the random baseline, particularly on English tasks. Overall, the XLM-R$_{\text{LARGE}}$ model performs better than the BART$_{\text{LARGE}}$ models in all tasks except SNIPS.

\paragraph{GPT vs. T5 models.}
In general, the GPT models outperform the T5 models in all language pairs and datasets in a head-to-head comparison: Both GPT-2$_{\text{LARGE}}$ and T5$_{\text{LARGE}}$ have a similar number of parameters (\texttt{0.8B}), but they have a significant performance difference. A similar pattern can also be observed on larger models, such as GPT$_\text{NEO}$ \texttt{2.7B} and T5$_{\text{3B}}$ \texttt{3B}. Although the T5 models perform worse than the GPT models, they do not have a maximum token size for the input, as the GPT models do, which is one of the advantages of using them. On the other hand, we find that changing the sample order tremendously affects the performance of the T5 models. As shown in Tables~\ref{monolingual-results} and~\ref{crosslingual-results}, the performance increases substantially when we sort the few-shot samples based on their label (i.e., first all positive and then all negative examples). Conversely, the GPT models suffer loss in performance. Thus, we can make the conclusion that changing the sample order may produce high variance in the results, as also shown in ~\cite{lu2021fantastically}.

% \begin{figure}[!th]
%   \centering
%   \includegraphics[width=\linewidth]{images/acc_snips_en_en.png} 
%   \caption{The accuracy of English SNIPS}
%   \label{fig:result-snips}
% \end{figure}

\begin{figure*}[!htb]
    \centering
    \begin{minipage}{.48\textwidth}
        \centering
        \includegraphics[width=0.99\linewidth]{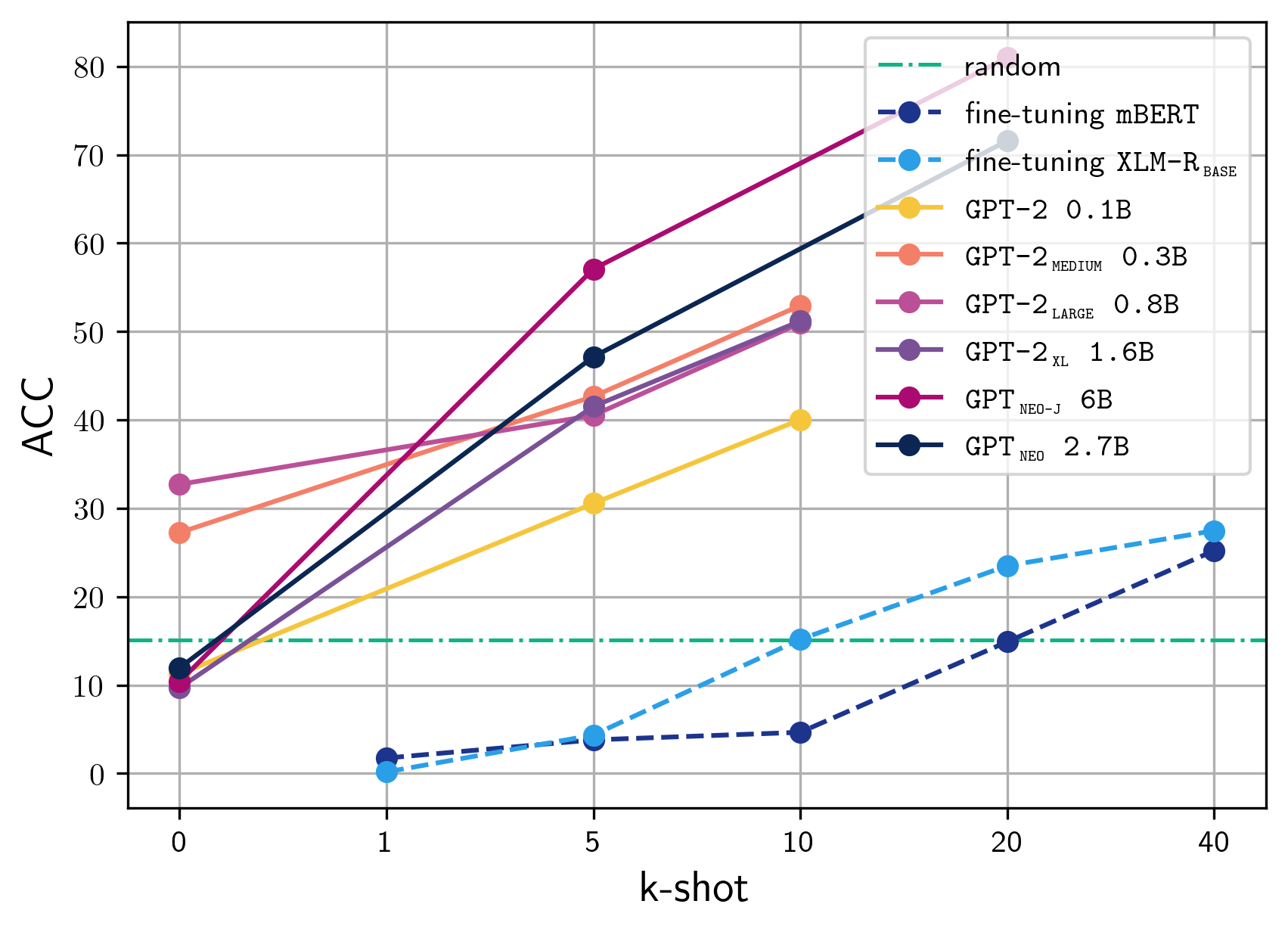}
        \caption{The results on German (de) MTOP dataset with GPT models.}
        \label{fig:gpt1}
    \end{minipage}
    \hspace{2mm}
    \begin{minipage}{0.48\textwidth}
        \centering
        \includegraphics[width=0.99\linewidth]{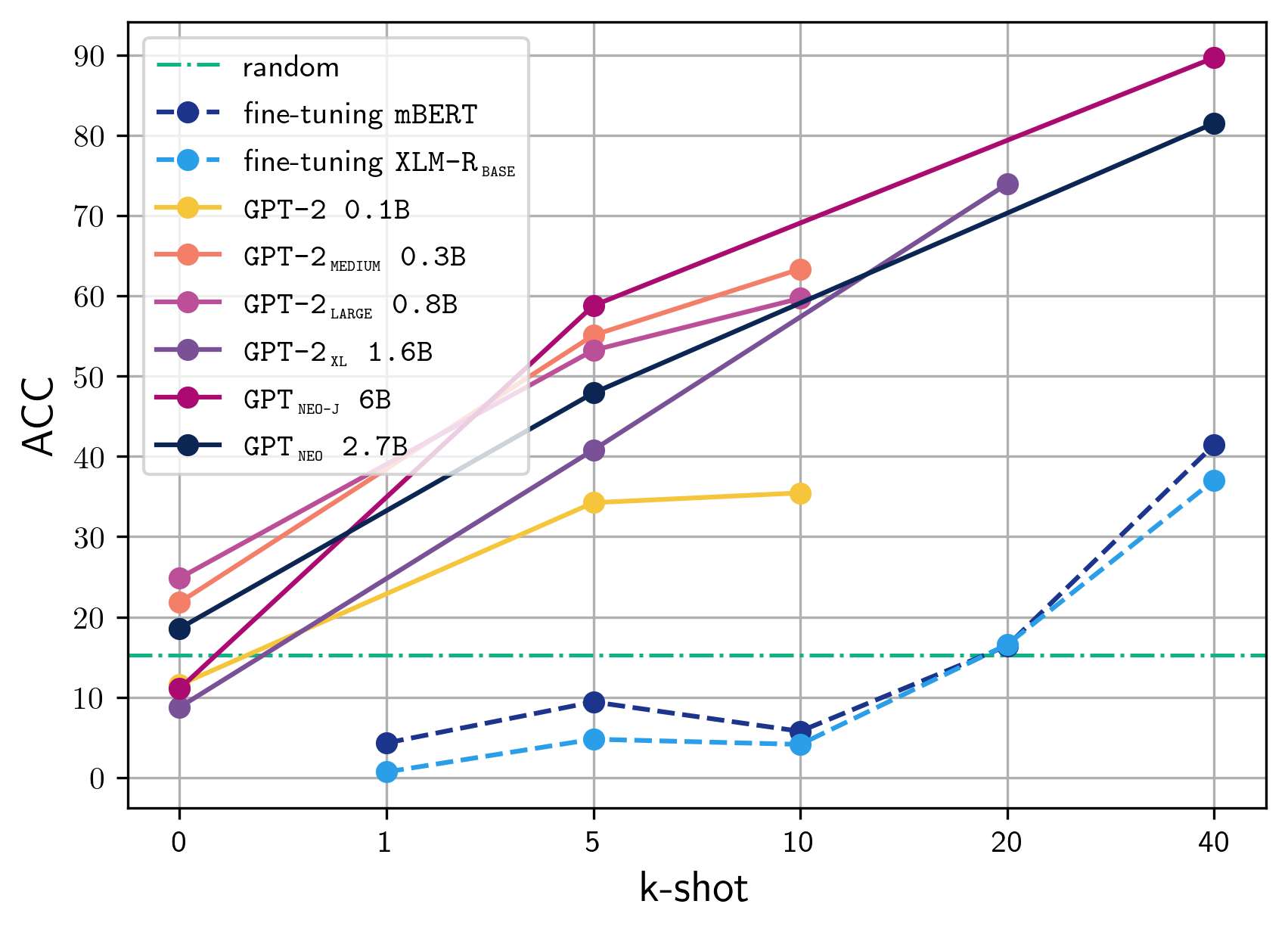}
        \caption{The results on English (en) MTOP dataset with GPT models.}
        \label{fig:gpt2}
    \end{minipage}
\end{figure*}
\begin{figure*}[!htb]
    \centering
    \begin{minipage}{.48\textwidth}
        \centering
        \includegraphics[width=0.99\linewidth]{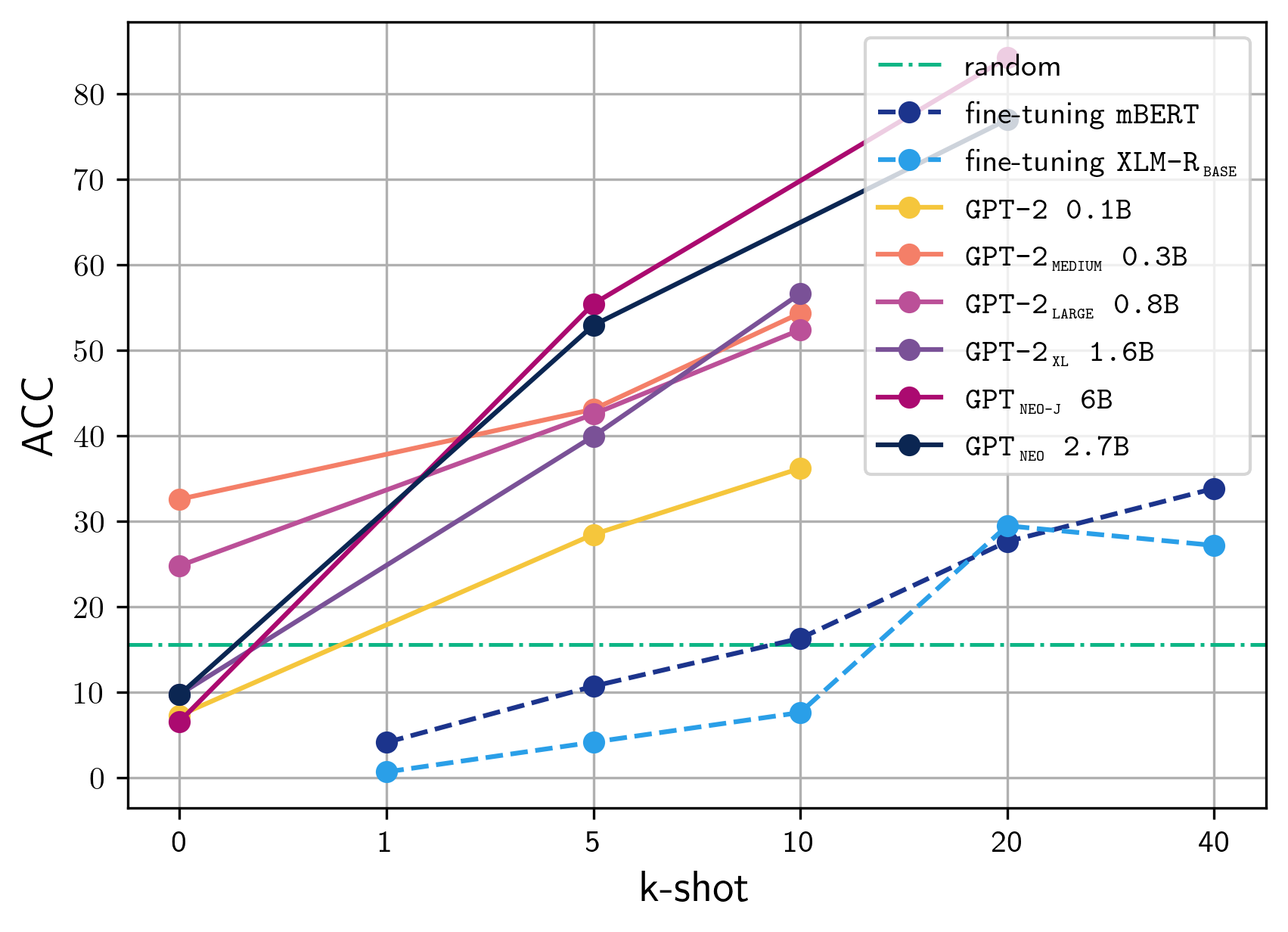}
        \caption{The results on Spanish (es) MTOP dataset with GPT models.}
        \label{fig:gpt3}
    \end{minipage}
    \hspace{2mm}
    \begin{minipage}{0.48\textwidth}
        \centering
        \includegraphics[width=0.99\linewidth]{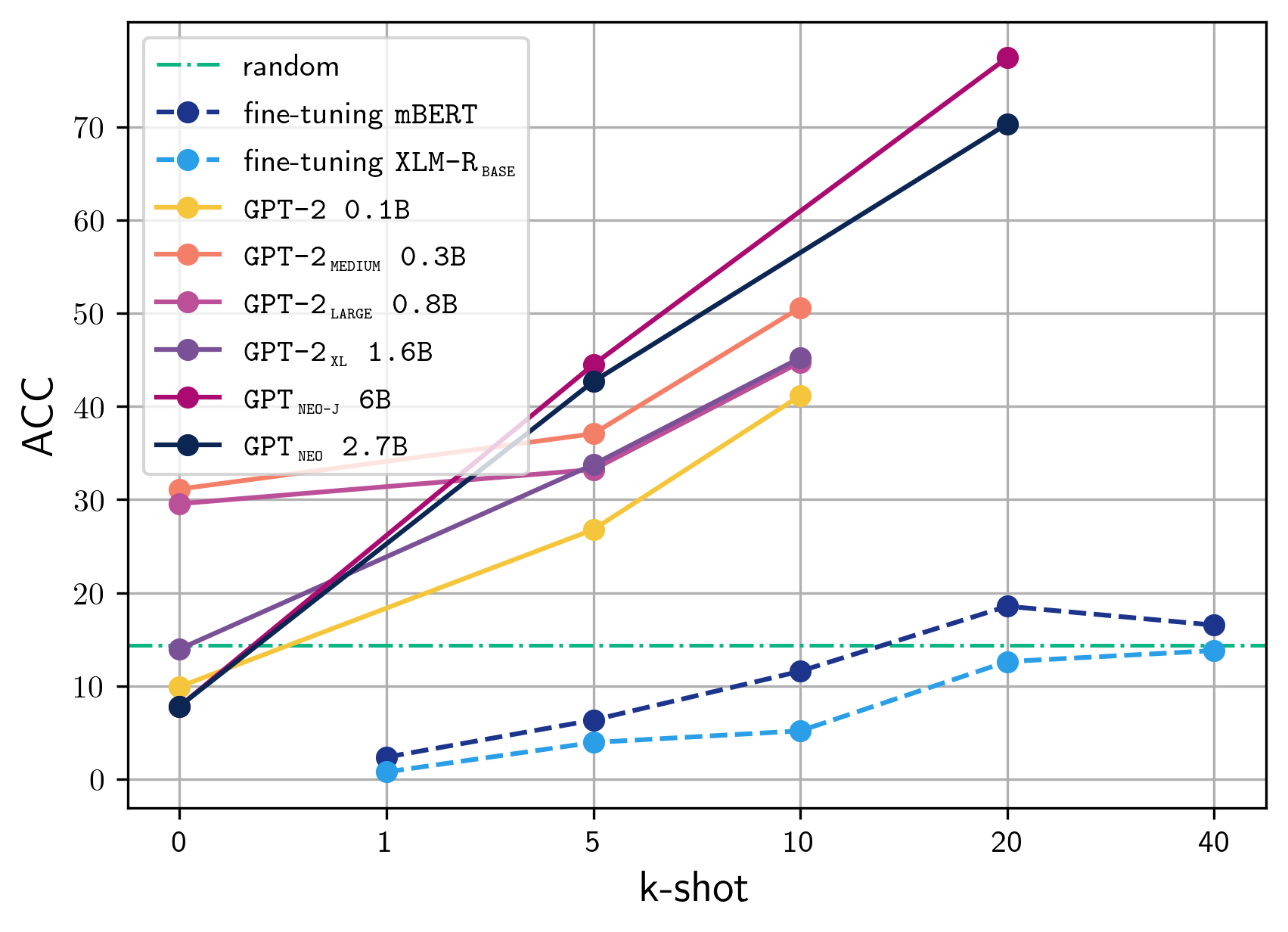}
        \caption{The results on French (fr) MTOP dataset with GPT models.}
        \label{fig:gpt4}
    \end{minipage}
\end{figure*}

\paragraph{Effectiveness on non-English languages.}
Based on the results, the performance of the models is lower in the non-English languages than in English. These results are expected since the pre-trained models are mostly trained on English data. However, the differences in performance are marginal. This finding may indicate that our few-shot learning method can be effectively utilized for languages that are in the same language family as English, such as French, German, and Spanish, but this will require further investigation in the future.

\paragraph{Cross-lingual results.}
Based on the results in Table~\ref{crosslingual-results}, we can see that the generative models are able to use the context from English to predict the sample in non-English languages. The cross-lingual setting is considered harder than the monolingual one since the models need to contextualize and understand the source and target languages to predict the test samples correctly. In general, the trend of the results in the cross-lingual setting is similar to the monolingual setting. In the MTOP dataset, we find that the models generally achieve higher performance for \textbf{en}$\rightarrow$\textbf{es} than for the other two target languages (\textbf{de} and \textbf{fr}). In MultiNLU, our GPT$_{\text{NEO-J}}$ closes the gap with the existing state-of-the-art baseline with fine-tuning from~\citet{liu2020attention} underperforming it only by a close margin of around 4.2\%, and the GPT$_{\text{NEO-J}}$ performance is only less than 3\% worse than that of the Translate-Train model. These results show a promising new direction in the zero-shot cross-lingual research that can be applied to other datasets and language pairs.

\begin{figure*}[!htb]
    \centering
    \begin{minipage}{.48\textwidth}
        \centering
        \includegraphics[width=0.99\linewidth]{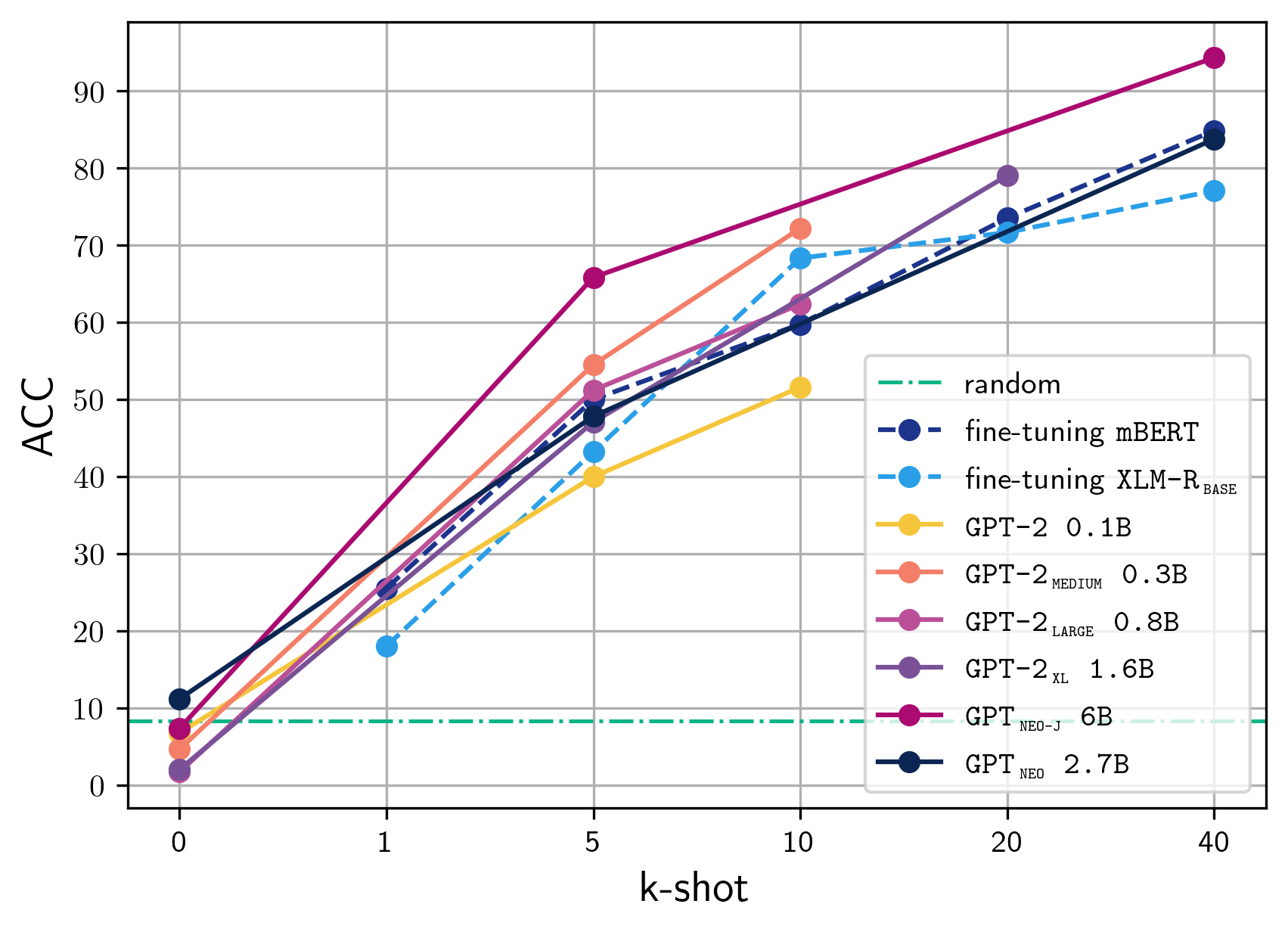}
        \caption{The results on English (en) multilingual NLU dataset with GPT models.}
        \label{fig:gpt5}
    \end{minipage}
    \hspace{2mm}
    \begin{minipage}{0.48\textwidth}
        \centering
        \includegraphics[width=0.99\linewidth]{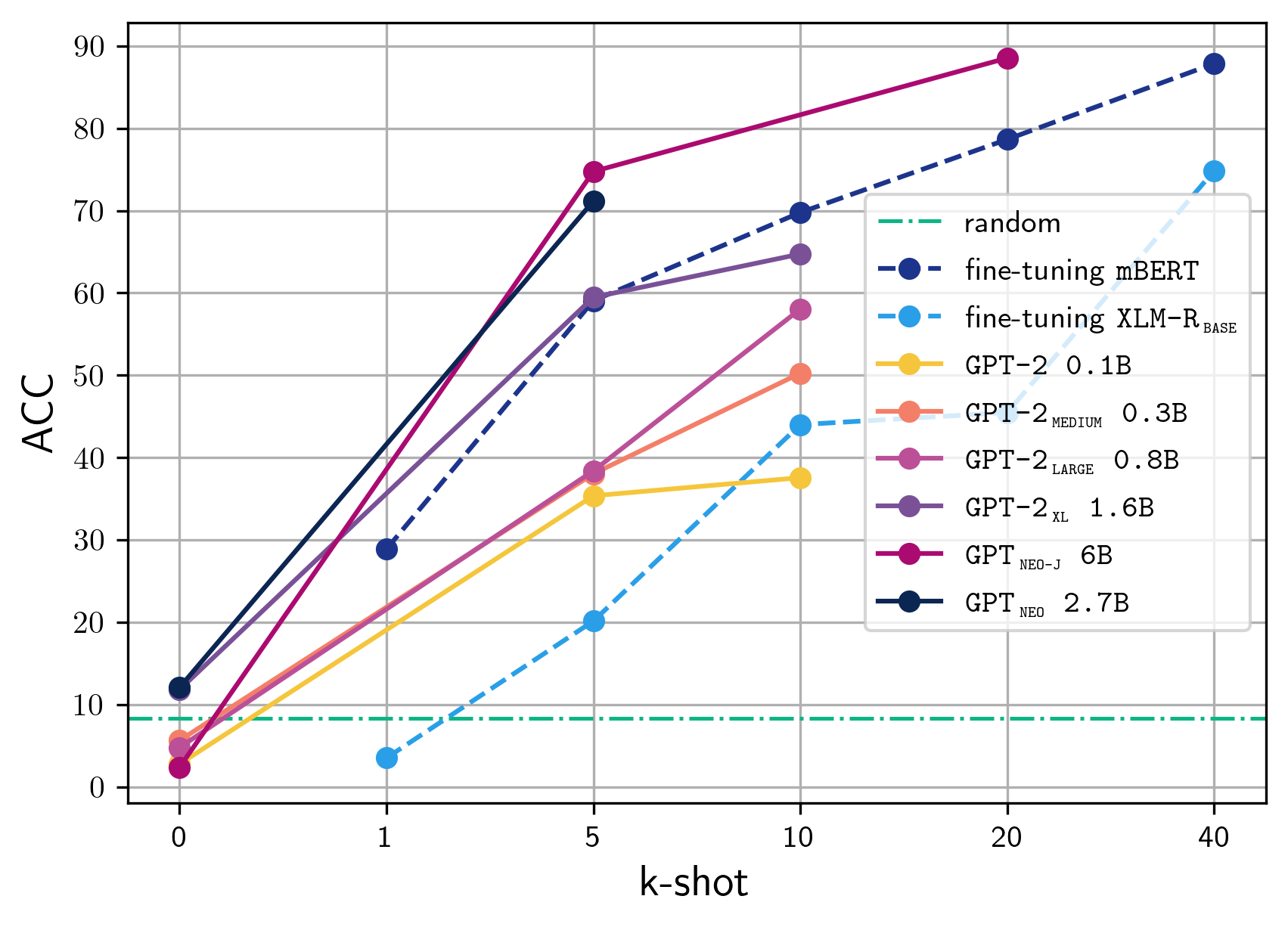}
        \caption{The results on Spanish (es) multilingual NLU dataset with GPT models.}
        \label{fig:gpt6}
    \end{minipage}
\end{figure*}

\subsection{Ablation Study}
To further understand how much data we need for the in-context learning, we conduct experiments with different numbers of few-shot samples, including zero-shot experiments on the MTOP and MultiNLU datasets. 

\paragraph{MTOP dataset.} Figures~\ref{fig:gpt1},~\ref{fig:gpt2},~\ref{fig:gpt3}, and~\ref{fig:gpt4} illustrate the results with different numbers of samples on the MTOP dataset in the monolingual setting. We show a different set of k-shot results for each model according to the maximum samples that can be used in the model as input. The results consistently improved as the number of shots increases. Interestingly, the QA style's zero-shot strategy can outperform random prediction only on two or three models in each language, and the others are worse. The fine-tuning results on MTOP are thus far worse than those of few-shot learning.

\paragraph{MultiNLU dataset.} Figures~\ref{fig:gpt5} and~\ref{fig:gpt6} illustrate the results with different numbers of samples on the MultiNLU dataset in the monolingual setting. The results on MultiNLU for the models with fine-tuning are closer to those of few-shot learning than those on the MTOP dataset. The reason may be the number of labels that the MTOP dataset has compared to MultiNLU. As a result, the zero-shot performance on the GPT models is sometimes worse than that of the random baseline. 

\section{Related Work}

\subsection{Few-shot In-Context Learning}
Recent work on few-shot in-context learning uses LMs to solve NLP tasks~\cite{petroni2019language,brown2020language,gao2020making,madotto2020language,zhao2021calibrate,schick2021s,lin2021leveraging}. In this approach, we select the appropriate prompts to trigger the LMs to behave so that they can predict the desired output~\cite{liu2021pre}. However, the prompts have to be engineered to allow the LM to generate a text appropriate to solve the task. 
Learning to calibrate the few-shot results is also essential to reduce the model's performance variance~\cite{zhao2021calibrate}, and the selection criteria in choosing the prompts are also important~\cite{perez2021true}. In another stream of work,~\citet{shin2020eliciting,li2021prefix} proposed an automated method to create prompts for a diverse set of tasks by gradient-based tuning instead of manually searching for a good prompt. Using such a method, may allow us to find an optimal prompt easier, it is very difficult to discover the optimal prompts for complicated natural language processing tasks, such as semantic parsing~\cite{liu2021pre}.

\subsection{Pre-trained Language Models}
Recent advances in pre-trained LMs have been focused on building pre-trained encoders, such as BERT~\cite{devlin2019bert}, RoBERTa~\cite{liu2019roberta}, ELMO~\cite{peters2018deep}, ULMFiT~\cite{howard-ruder-2018-universal}, ELECTRA~\cite{clark2019electra}, XLM~\cite{conneau2019cross}, and XLM-R~\cite{conneau2020unsupervised,goyal2021larger}, decoder-only models, such as GPT models~\cite{radford2019language,brown2020language} and encoder-decoder models, such as T5~\cite{raffel2020exploring},  BART~\cite{lewis2020bart}, and their multilingual versions, mT5~\cite{xue2021mt5} and mBART~\cite{liu2020multilingual}.

Pre-trained encoders have been used to improve the contextualized representations of multilingual systems in various NLP tasks, for example, dialogue systems~\cite{liu2020attention, liu-etal-2021-x2parser, li2021mtop}, code-switching sequence labeling~\cite{Aguilar2020Char2SubwordET,winata2021multilingual,winata2021thesis}, and multilingual speech recognition~\cite{datta2020language,winata2020adapt}. Meanwhile, the pre-trained encoder-decoder models, have been used for various sequence generation tasks, such as summarization~\cite{raffel2020exploring}, conversational agents~\cite{lin2020mintl,lin2020xpersona,madotto2020learning,wu2020probing,hosseini2020simple,lin2021bitod}, and knowledge grounding~\cite{chen-etal-2020-kgpt,zhao-etal-2020-knowledge-grounded}.

\section{Conclusion}
This paper demonstrates the multilingual skills of pre-trained LMs, GPT and T5, in conducting in-context learning without parameter updates. This work is our initial attempt to show the effectiveness of in-context learning in the multilingual and cross-lingual setting. It covers four different languages and explores the possibility of conducting efficient inference on low-resource tasks. We find that LMs can predict samples correctly, significantly better than random prediction, in cross-lingual tasks with no training examples of the target languages. We would like to further investigate the applicability of this method to other tasks and languages in future work.

\section*{Acknowledgment}
We want to thank Bryan Wilie and Samuel Cahyawijaya for their support in accessing the cloud service. We also sincerely thank Zihan Liu and ML Collective members for helping with the discussion about this project.

\bibliography{anthology,custom}
\bibliographystyle{acl_natbib}

\appendix
\section{Full k-shot Results}
\label{sec:appendix}

This appendix shows the results on few-shot monolingual and cross-lingual settings on SNIPS, MTOP, and multilingual NLU datasets over a different number of samples.

\begin{figure*}[!htb]
    \centering
    \begin{minipage}{.48\textwidth}
        \centering
        \includegraphics[width=0.99\linewidth]{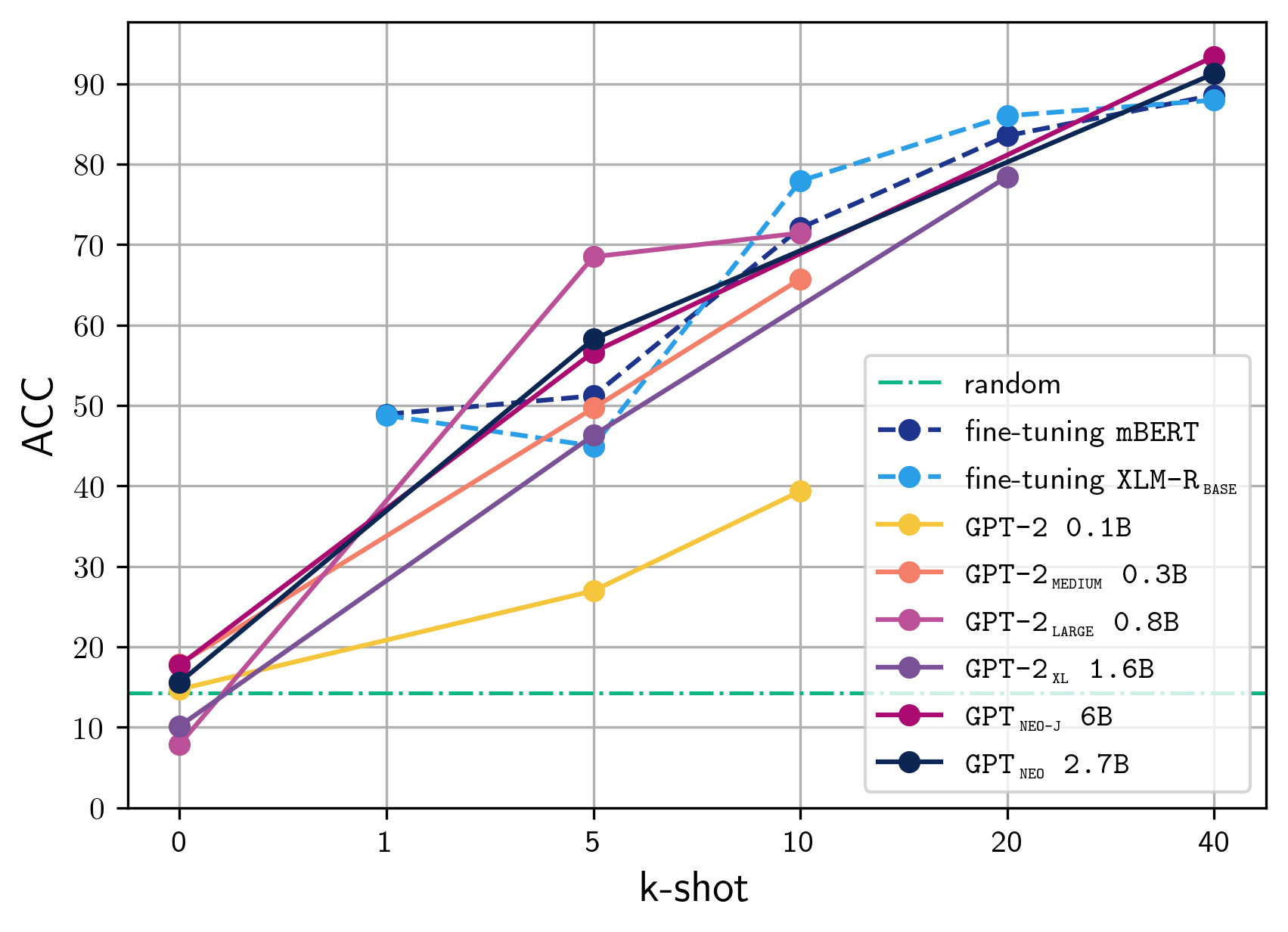}
        \caption{The acc results on English (en) SNIPS with GPT models.}
        \label{fig:snips-acc}
    \end{minipage}
    \hspace{2mm}
    \begin{minipage}{0.48\textwidth}
        \centering
        \includegraphics[width=0.99\linewidth]{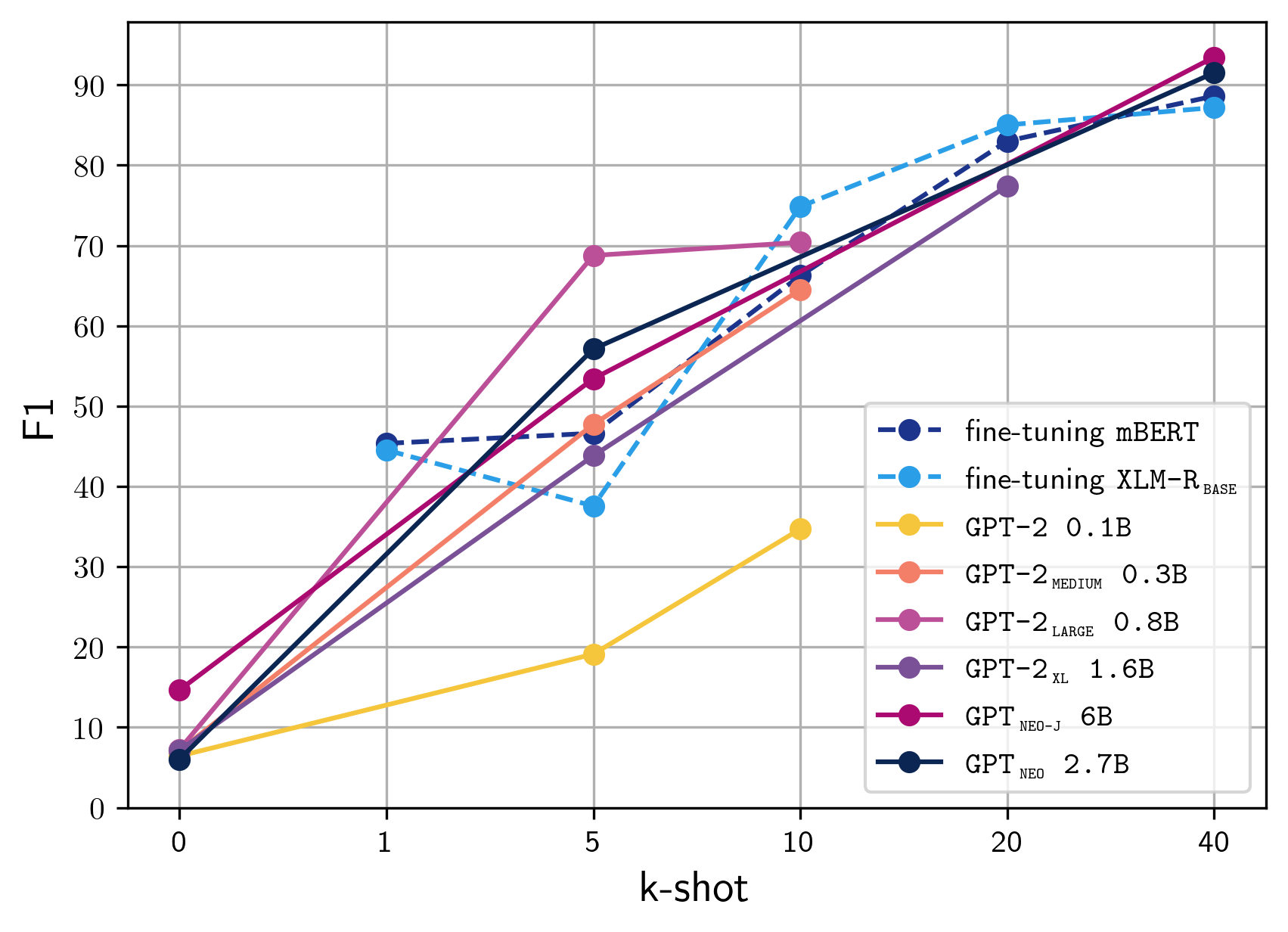}
        \caption{The f1 results on English (en) SNIPS with GPT models.}
        \label{fig:snips-f1}
    \end{minipage}
\end{figure*}
\begin{figure*}[!htb]
    \centering
    \begin{minipage}{.48\textwidth}
        \centering
        \includegraphics[width=0.99\linewidth]{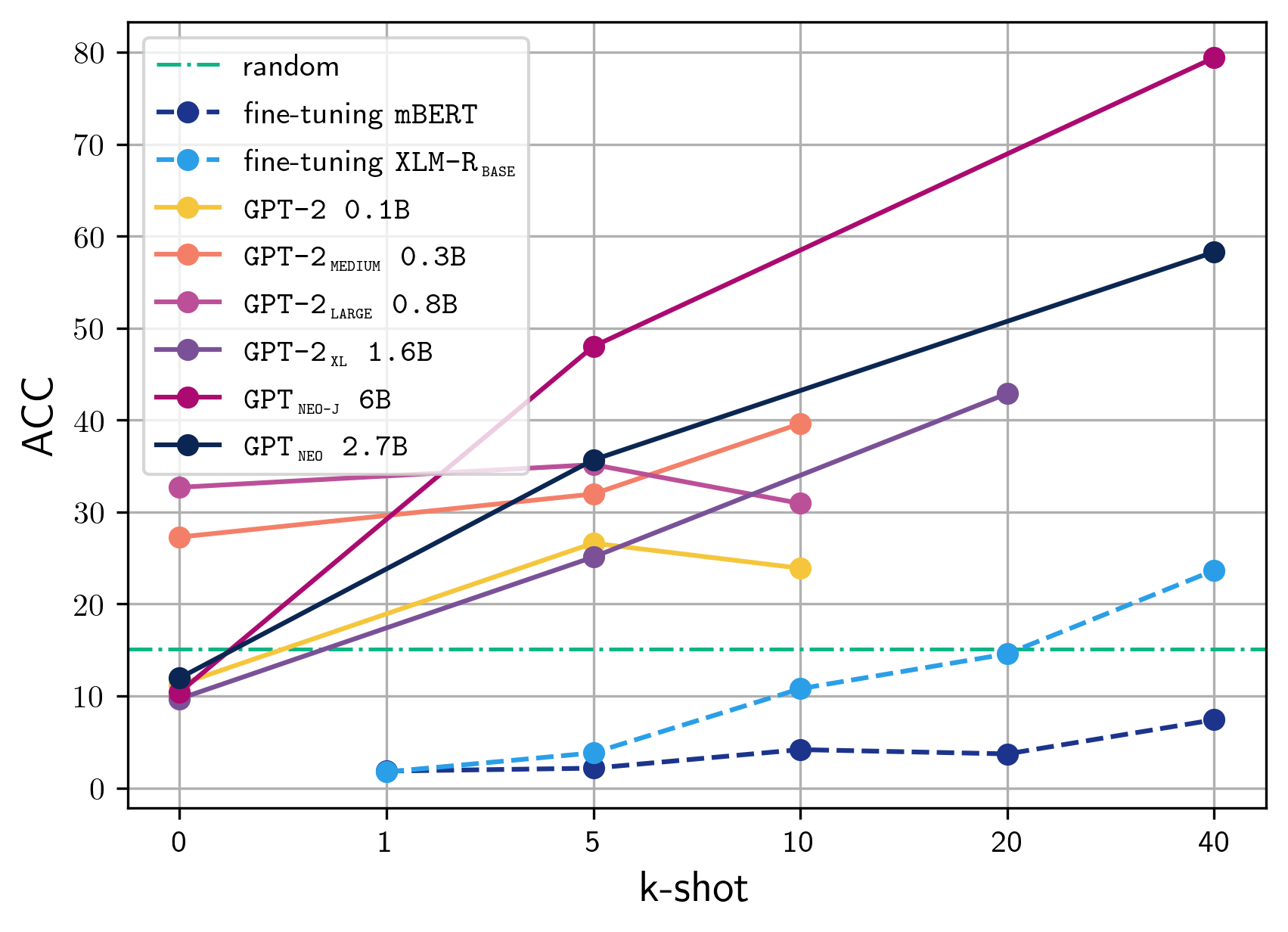}
        \caption{The acc results on the cross-lingual setting, English-German (de) MTOP dataset with GPT models.}
        \label{fig:mtop-en-de-acc}
    \end{minipage}
    \hspace{2mm}
    \begin{minipage}{.48\textwidth}
        \centering
        \includegraphics[width=0.99\linewidth]{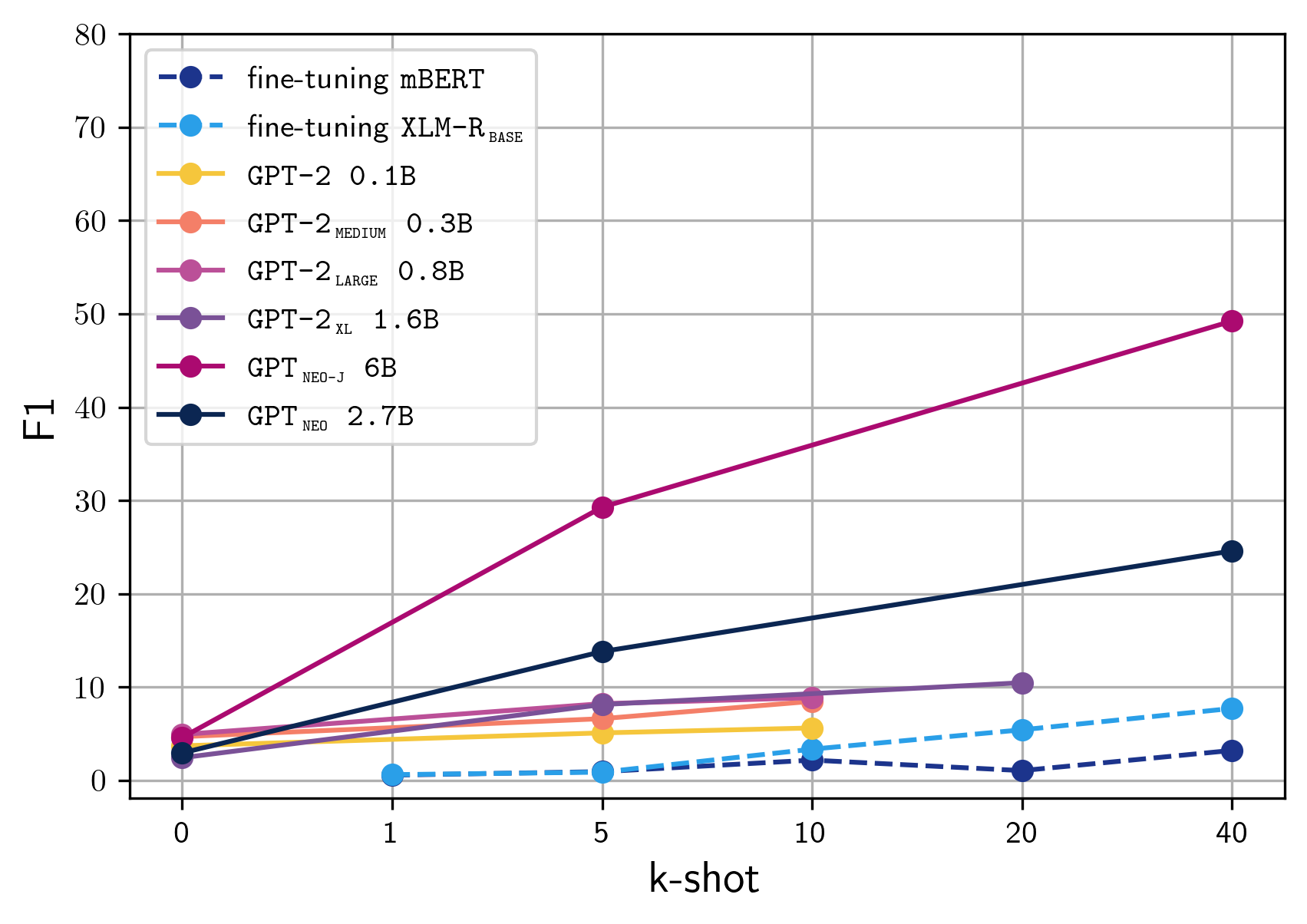}
        \caption{The f1 results on the cross-lingual setting, English-German (de) MTOP dataset with GPT models.}
        \label{fig:mtop-en-de-f1}
    \end{minipage}
\end{figure*}
\begin{figure*}[!htb]
    \centering
    \begin{minipage}{.48\textwidth}
        \centering
        \includegraphics[width=0.99\linewidth]{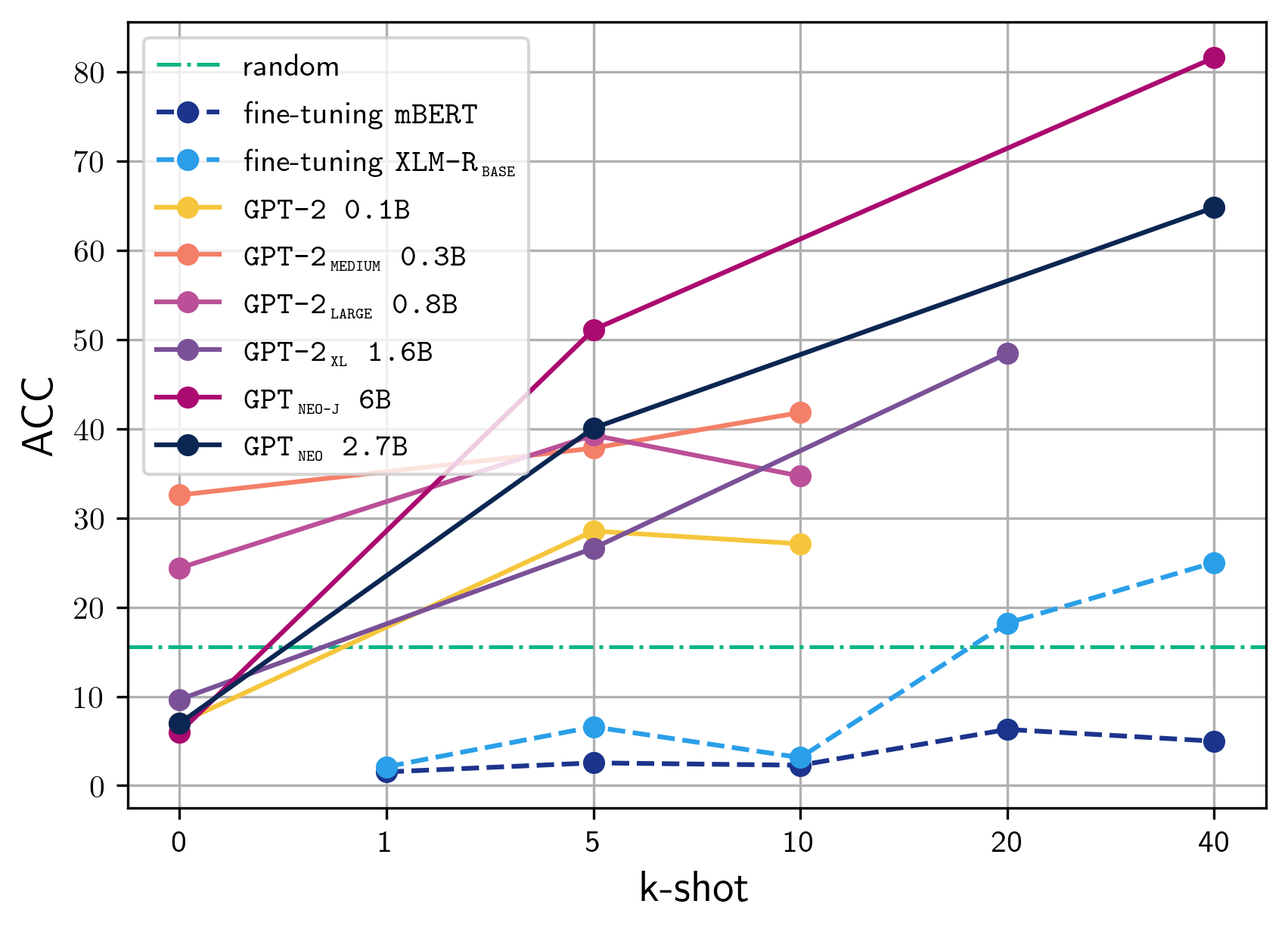}
        \caption{The acc results on the cross-lingual setting, English-Spanish (es) MTOP dataset with GPT models.}
        \label{fig:mtop-en-es-acc}
    \end{minipage}
    \hspace{2mm}
    \begin{minipage}{.48\textwidth}
        \centering
        \includegraphics[width=0.99\linewidth]{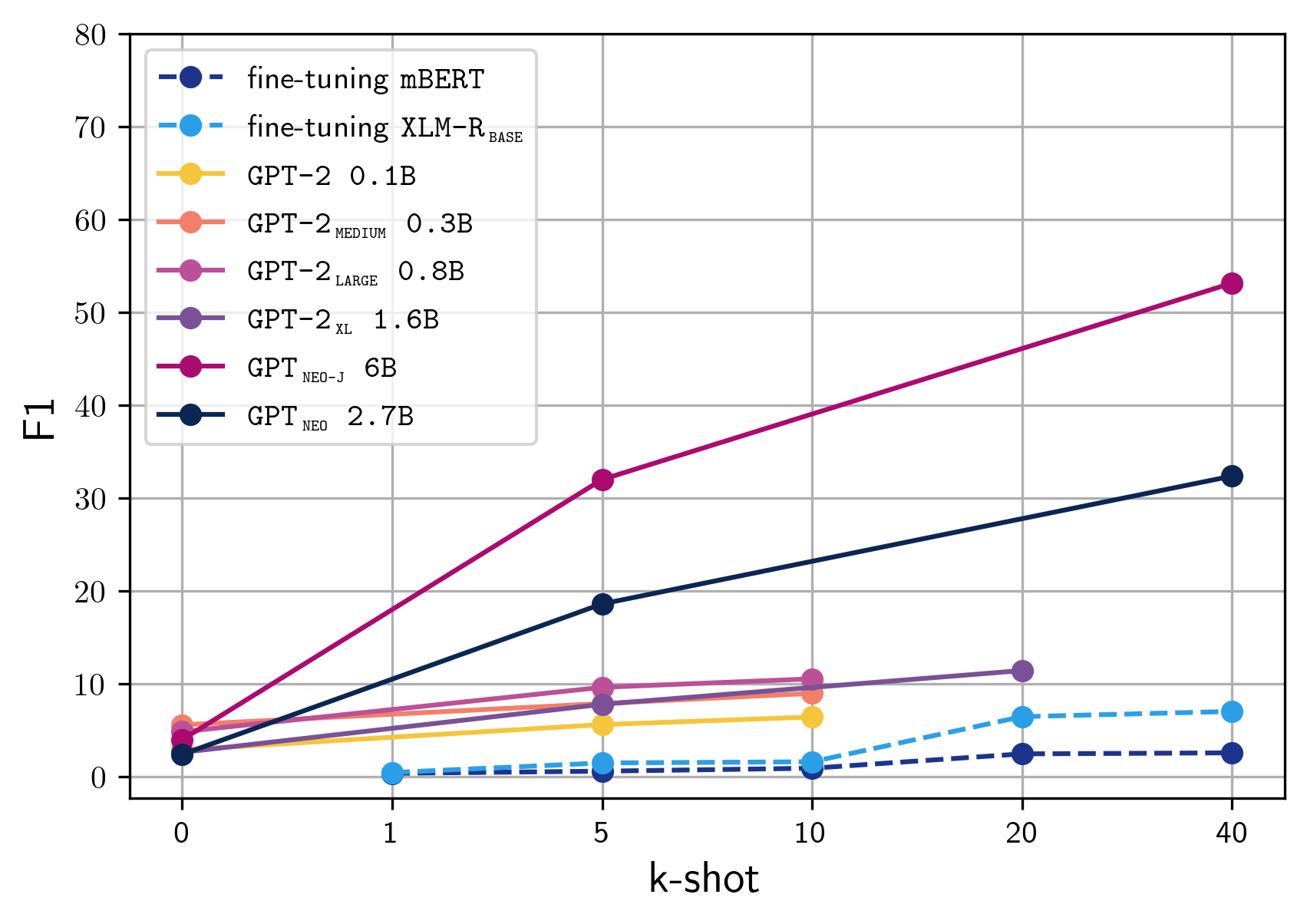}
        \caption{The f1 results on the cross-lingual setting, English-Spanish (es) MTOP dataset with GPT models.}
        \label{fig:mtop-en-es-f1}
    \end{minipage}
\end{figure*}
\begin{figure*}[!htb]
    \centering
    \begin{minipage}{.48\textwidth}
        \centering
        \includegraphics[width=0.99\linewidth]{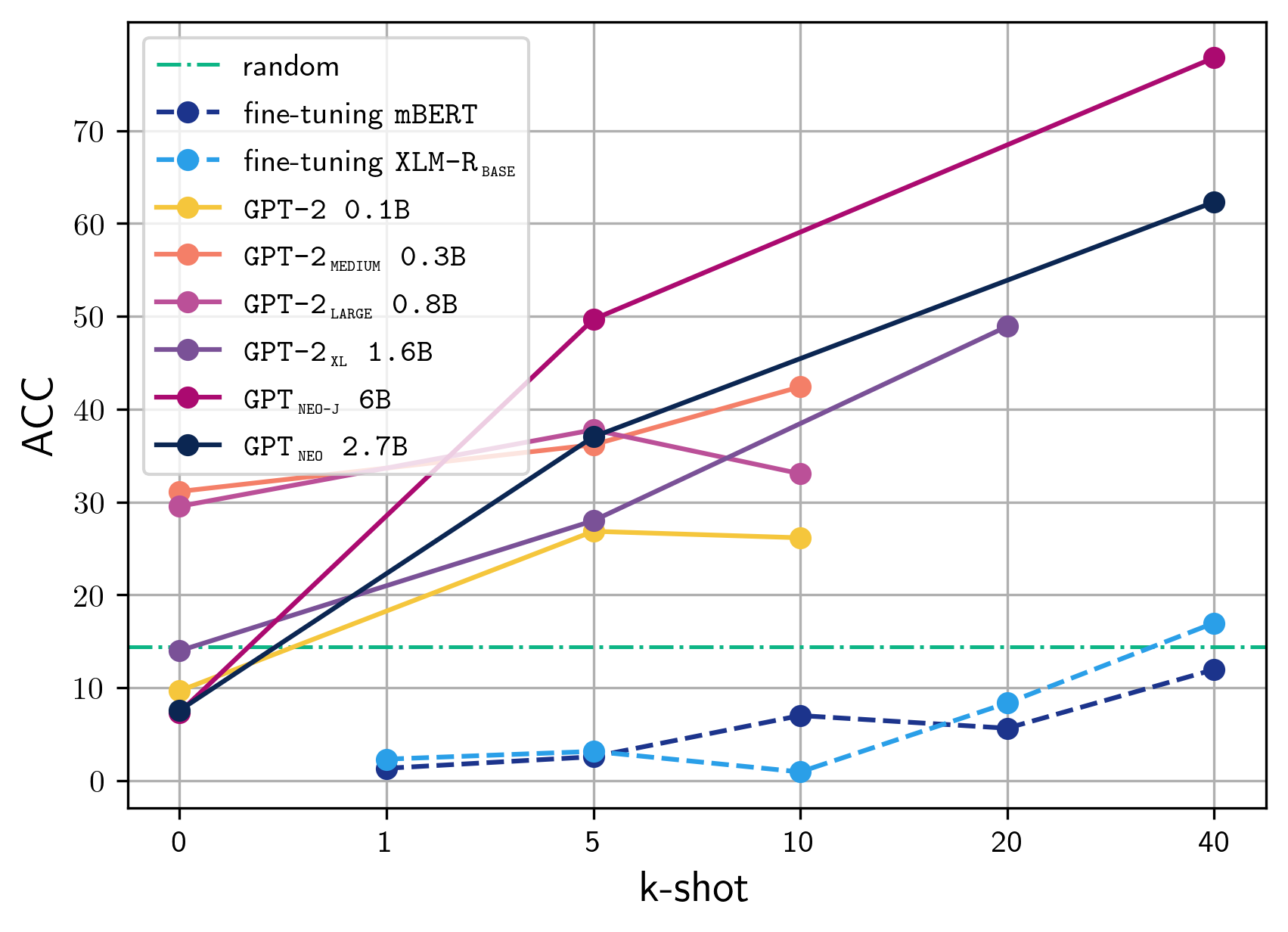}
        \caption{The acc results on the cross-lingual setting, English-French (fr) MTOP dataset with GPT models.}
        \label{fig:mtop-en-fr-acc}
    \end{minipage}
    \hspace{2mm}
    \begin{minipage}{.48\textwidth}
        \centering
        \includegraphics[width=0.99\linewidth]{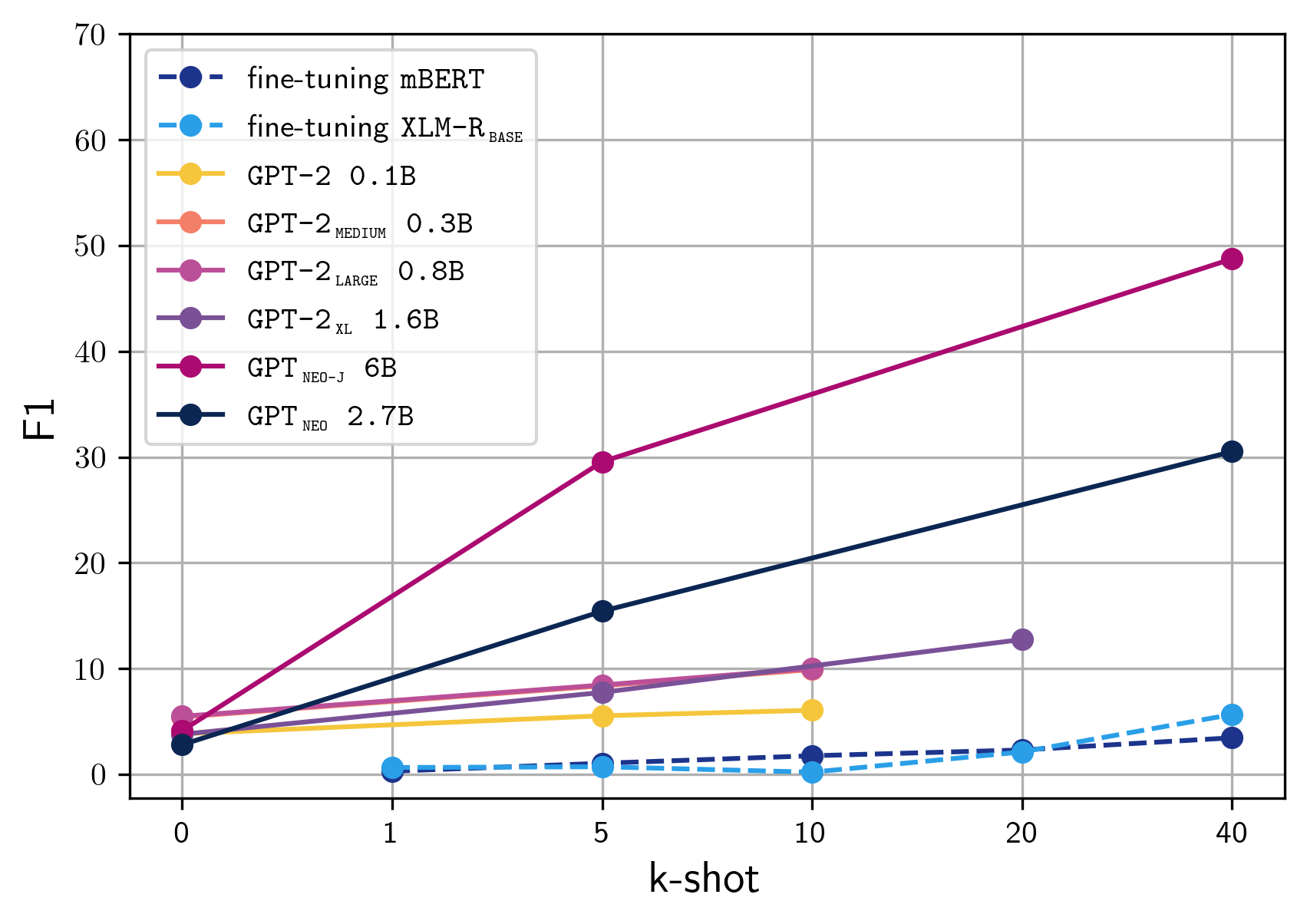}
        \caption{The f1 results on the cross-lingual setting, English-French (fr) MTOP dataset with GPT models.}
        \label{fig:mtop-en-fr-f1}
    \end{minipage}
\end{figure*}
\begin{figure*}[!htb]
    \centering
    \begin{minipage}{.48\textwidth}
        \centering
        \includegraphics[width=0.99\linewidth]{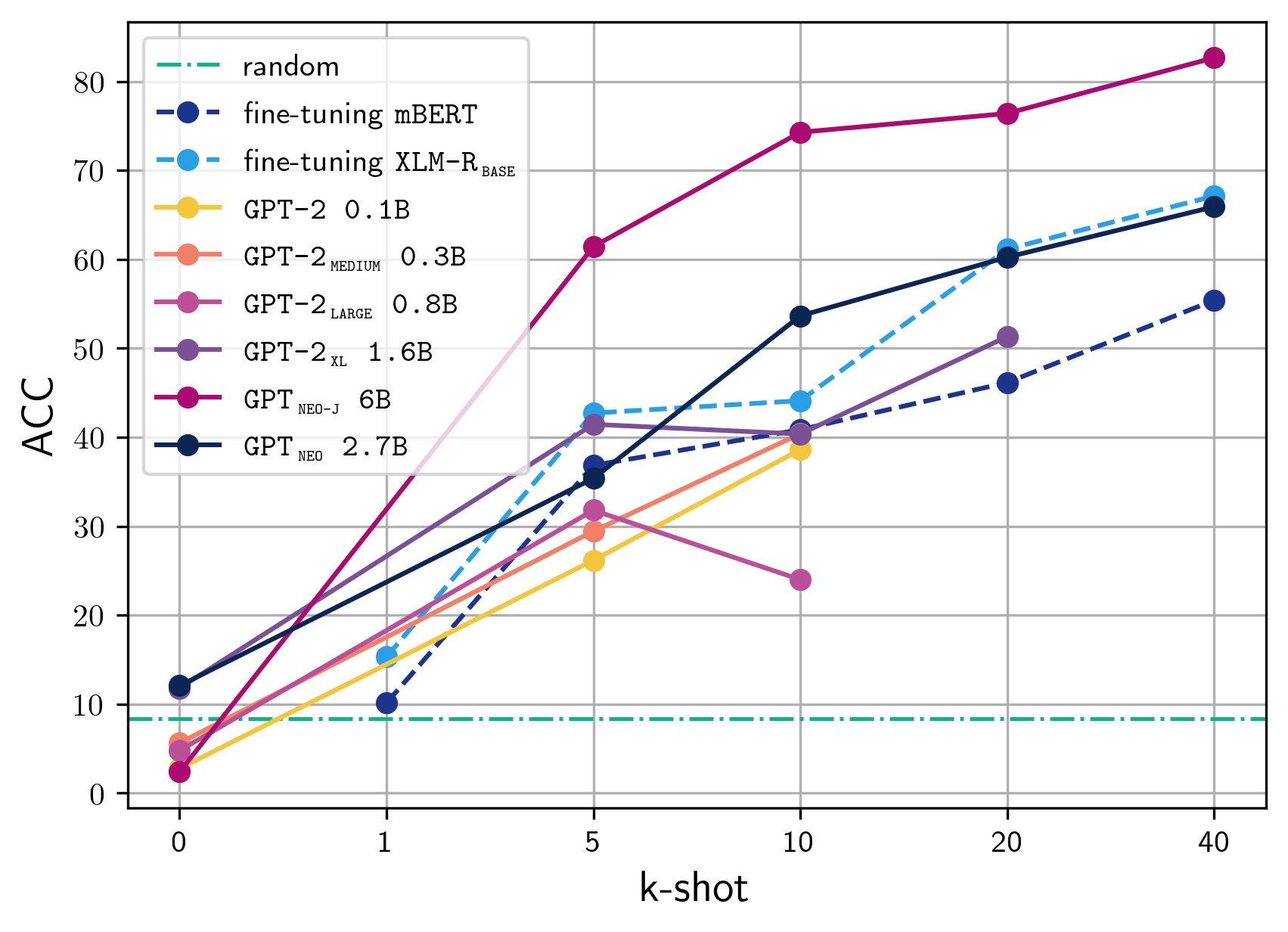}
        \caption{The acc results on the cross-lingual setting, English-Spanish (es) multilingual NLU dataset with GPT models.}
        \label{fig:multi-nlu-en-es-acc}
    \end{minipage}
    \hspace{2mm}
   \begin{minipage}{.48\textwidth}
        \centering
        \includegraphics[width=0.99\linewidth]{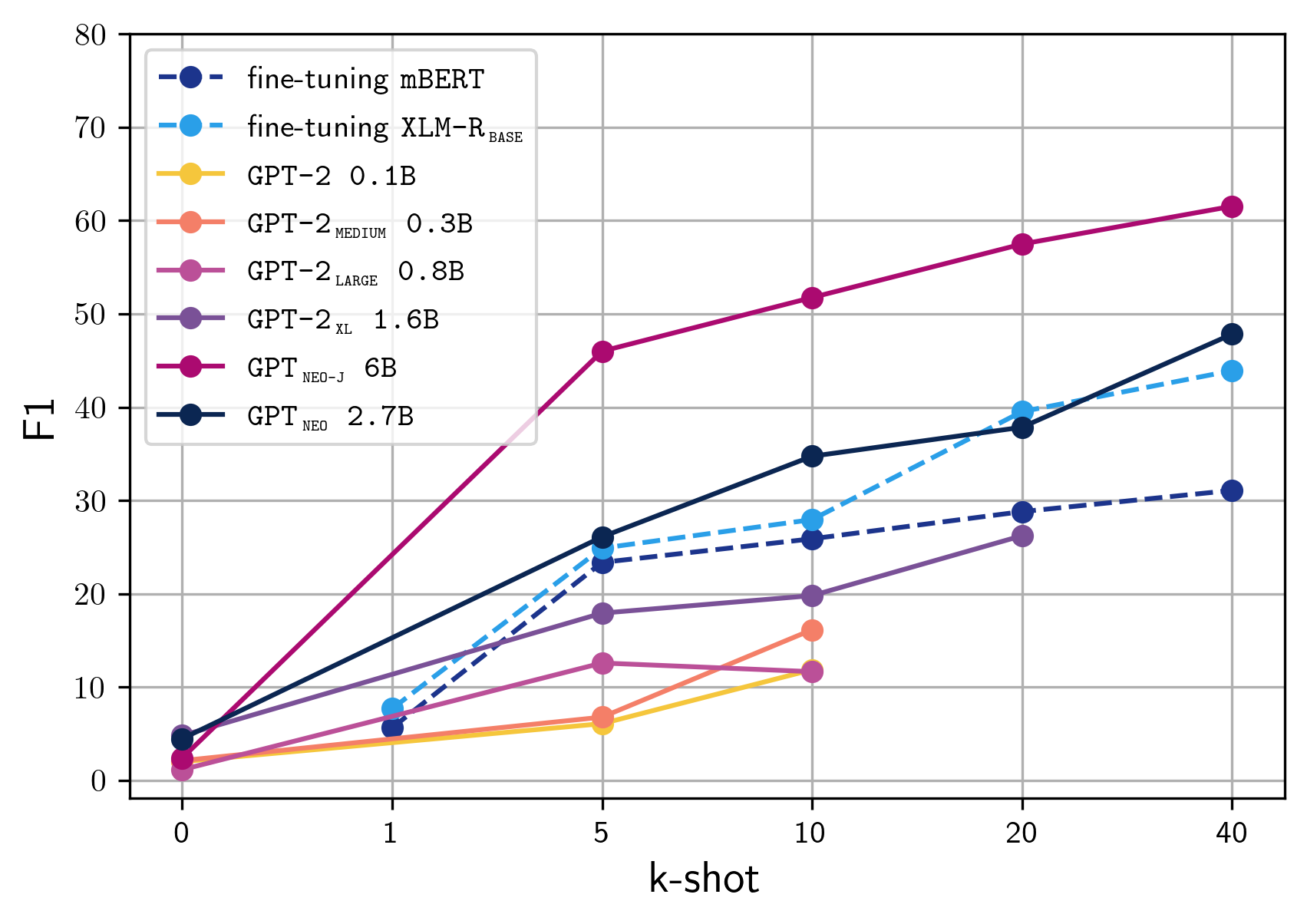}
        \caption{The f1 results on the cross-lingual setting, English-Spanish (es) multilingual NLU dataset with GPT models.}
        \label{fig:multi-nlu-en-es-f1}
    \end{minipage}
\end{figure*}

\end{document}